\documentclass[twocolumn,letterpaper]{IEEEAerospaceCLS}
\usepackage[T1]{fontenc}
\usepackage{graphicx}
\usepackage{amsmath,amssymb}
\usepackage{booktabs,tabularx,array}
\usepackage{microtype}
\usepackage{placeins}
\usepackage{url}
\newcounter{subsubsection}[subsection]
\usepackage[colorlinks=true,urlcolor=blue,linkcolor=black,citecolor=black]{hyperref}
\graphicspath{{figures/print/}{figures/}}

\hypersetup{pdftitle={What Frame-Level Labels Can and Cannot Do for Small-UAV Point Detection in Thermal Video},pdfauthor={Wonbin Son, Gyumun Choi, Junil Seo, Hyungjoon Kim},pdfsubject={Author-prepared revised preprint submitted to IEEE for possible publication}}

\makeatletter
\newcommand{\bstctlcite}[1]{\@bsphack\if@filesw\immediate\write\@auxout{\string\citation{#1}}\fi\@esphack}
\makeatother

\begin{document}
\bstctlcite{IEEEauthorcontrol}
\raggedbottom
\title{What Frame-Level Labels Can and Cannot Do for Small-UAV Point Detection in Thermal Video}
\author{
\begin{tabular*}{6.8in}[b]{@{\extracolsep{\fill}}cccc@{}}
Wonbin Son\textsuperscript{*} & Gyumun Choi & Junil Seo & Hyungjoon Kim\textsuperscript{\ensuremath{\dagger}}\\
diwjidghk78@gmail.com & cgymun@gmail.com & junil0106@gmail.com & hyungjoon@changwon.ac.kr
\end{tabular*}\\[6pt]
{\small\textsuperscript{*}First author \qquad \textsuperscript{\ensuremath{\dagger}}Corresponding author}
\thanks{\footnotesize This work has been submitted to the IEEE for possible publication. Copyright may be transferred without notice, after which this version may no longer be accessible.}
}
\maketitle
\thispagestyle{plain}
\pagestyle{plain}

\begin{abstract}\bfseries

The growing use of unmanned aerial vehicles (UAVs) has increased the importance of image-based UAV detection. Learning-based detectors used for this task are trained on imagery and annotations, with the annotation type determining the information available during training. Among the available approaches, we focus on learning localization from frame-level target presence/absence labels when sensor or scene changes make obtaining spatial annotations for additional training burdensome. In this context, we analyze the detection capability, learning and detection behavior, and potential applications of an existing architecture for point detection of small UAVs, trained with presence/absence labels and requiring no external detector. The architecture freezes spatial features learned through classification and trains a readout with the same frame labels to produce spatial score maps and point detections. On two thermal infrared datasets, CST Anti-UAV and Anti-UAV410, we evaluate localization hit rates and detection rates under false-alarm constraints, analyze the effects of training stages, label allocation, synthesis, and model configuration, and compare with bounding-box detectors. We additionally explore potential applications through a supplementary analysis on Airborne Object Tracking (AOT), training separate models on its visible-light imagery using the dataset's own frame labels. Analysis of training stages suggested that classification training strengthens target-related spatial responses and that readout training helps extract them consistently. Distributing similar label counts across more videos yielded higher localization hit rates in our comparisons, while synthesis effects varied by dataset and evaluation criterion. Higher localization hit rates did not always lead to better detection under false-alarm constraints, and localization failures remained when target signals were weak relative to background variation and under cross-dataset transfer. These findings provide insights and guidance on label allocation, spatial representations and readouts, synthesis, and false-alarm control, considering annotation procedures, required localization precision, and stability across training runs.

\end{abstract}

\tableofcontents

\section{Introduction}

\label{sec:1}

\subsection{1.1\quad Background and Problem Statement}

\label{sec:1.1}

Unmanned aerial vehicles (UAVs) are used in applications including imaging, transportation, and surveillance. As their use expands, so does the need to monitor surrounding airspace and understand UAV activity across diverse operating environments. A basic task in this setting is to determine whether a UAV is present in the monitored area and locate it. In low-light conditions, such as nighttime, visible-light imagery alone may be insufficient for observing targets, motivating the use of thermal infrared imagery for UAV detection and tracking. Even in thermal imagery, however, reliable localization remains difficult when targets appear small or are hard to distinguish from the background \cite{xie2025cst}.

Another challenge is that changes in sensors or observation scenes can degrade an existing detector's performance, potentially requiring additional training on data from the intended deployment environment. Annotating new imagery can then become a burden, making the form of training labels an important choice. Bounding boxes (bboxes) provide target locations and sizes for training. In contrast, frame-level presence/absence labels indicate only whether a target is present in each frame. If localization can also be learned from these labels, specifying target locations during annotation can be replaced by judging target presence or absence. In video, frame labels can be assigned to verified intervals of UAV presence or absence by marking their temporal boundaries \cite{dutta2019via}. This can reduce repeated label entry when the presence state persists across consecutive frames.

Although presence/absence labels do not directly specify target locations, prior work has shown that networks trained only with image-level classification labels can produce spatial responses associated with the objects of interest \cite{zhou2016cam}. In the architecture studied here, a classifier is first trained to distinguish UAV presence from absence in each frame. A small output module, termed a readout, is then trained with the same presence/absence labels to convert the classifier's learned spatial features into scores at each location. At inference, locations with strong responses in the resulting score map serve as target candidates, yielding point detections from a single frame without an external detector.

We examine this architecture as one option for detecting small UAVs. We analyze its detection capability and learning and detection behavior, and consider its potential for further extension and application based on observed strengths and limitations. Through this analysis, we aim to provide insights and guidance for future architectural improvements and practical use.

\subsection{1.2\quad Related Work and Positioning of This Study}

\label{sec:1.2}

Several studies have investigated obtaining localization information from classifiers trained with image-level labels. Oquab et al. predicted object locations by directly reading the maximum-response locations in score maps learned with classification supervision, while the class activation mapping (CAM) approach of Zhou et al. constructed localization responses using global average pooling and classification weights \cite{oquab2015,zhou2016cam}. WILDCAT addressed classification, pointwise localization, and segmentation using image-level labels \cite{durand2017}. Subsequent weakly supervised object localization (WSOL) methods have explored learning from complementary regions and shallow features \cite{zhang2018acol,wei2021spol}, foreground prediction maps \cite{wu2022bas,li2026fgca}, and Transformer-based localization \cite{gao2021tscam,bai2022scm,wu2023sat}.

Closely related architectures have also been studied. Vardazaryan et al. used ResNet-18, a 1\ensuremath{\times}1 convolution, and spatial pooling, and directly output the maximum-response location of a map trained with surgical-tool presence/absence labels \cite{vardazaryan2018}. We use a simple classification backbone and readout as a common reference for controlled comparisons of data composition, supervision, spatial resolution, and training stages in thermal UAV point detection.

Prior work on infrared target detection has also explored image-level labels with CAM-based feature selection \cite{mccurley2022}, single-point supervision (LESPS) \cite{ying2023}, and target-count annotations with multi-frame input (WeCoL) \cite{duan2025}. These methods differ in annotation requirements and temporal inputs; this study focuses on frame-level binary labels and single-frame inference.

Comparisons of weakly supervised methods depend not only on the model but also on annotation use at different stages and the evaluation criteria. Prior evaluations have shown that spatial annotations used for model selection and threshold estimation can affect comparisons between WSOL methods \cite{choe2020,murtaza2025}. We distinguish annotations used for training, checkpoint selection, and calibration, and evaluate both localization hit rates and detection rates under false-alarm constraints. We also compare against bounding-box alternatives, including YOLO26n/s, to examine the applicability of the frame-label architecture in relation to performance requirements \cite{jocher2026}.

\subsection{1.3\quad Research Questions and Contributions}

\label{sec:1.3}

This study asks how well a direct point detection architecture trained with frame labels detects targets, which choices it is sensitive to, and under what conditions it falls short. To address these questions, we evaluate models trained with each dataset's own labels and cross-dataset transfer on two thermal infrared UAV datasets, CST Anti-UAV \cite{xie2025cst} and Anti-UAV410 \cite{huang2024}. We also analyze the learning of spatial responses and the effects of data and model configurations.

We additionally explore applicability beyond thermal imagery through a supplementary analysis that applies the same architecture to visible-light grayscale images from Airborne Object Tracking (AOT) \cite{amazonAOT}. Models are trained with helicopter presence/absence labels, and we examine localization performance on validation data and variability across training runs as the set of scenes is expanded.

The main contributions are as follows.

\begin{enumerate}

\item \textbf{Comparative evaluation of detection capability and limitations.} We compare frame-label point detectors with bounding-box-supervised heads on the same backbone, either frozen or fine-tuned, and with separate bounding-box detectors. This comparison shows why localization hit rates and detection under false-alarm constraints need to be evaluated separately.

\item \textbf{Analysis of characteristics arising from training, data, and model choices.} Controlled comparisons of labels and training stages examine spatial responses during early classification and the role of readouts on frozen features. We also assess how label allocation, synthesis, architecture, and inference configuration affect detection performance, training stability, and computational cost.

\item \textbf{Insights and application guidance from observations.} We analyze failures associated with low signal-to-clutter ratio (SCR), target size, scene composition, and cross-dataset transfer, deriving guidance on label allocation, model selection, synthesis, and false-alarm control. This guidance considers labels used for checkpoint selection and calibration, as well as variation across training runs, within explicitly stated applicability limits.

\end{enumerate}

These analyses and observations provide a basis for understanding the architecture's potential and characteristics. We hope that the resulting insights and guidance will support research on further improvements to frame-label detection architectures and their extension to other application domains.


\section{Study Setup: Task, Pipeline, and Evaluation}

\label{sec:2}

\subsection{2.1\quad Task and Scope of Supervision}

\label{sec:2.1}

Our main analysis examines a procedure that directly uses spatial responses learned from frame-level presence/absence labels for UAV point detection in thermal infrared imagery. The training data consist of frames $x_i$ and binary labels $y_i\in\{0,1\}$. These labels indicate the presence of the UAV designated as the tracking target in each dataset. The detector takes a single frame as input, produces a spatial response map, and extracts candidate locations and their scores. No preceding or subsequent frames or initial target location are provided at inference. We evaluate localization near the target center and detection under false-alarm constraints; bounding-box size estimation and temporal tracking are outside the scope of this evaluation.

Here, \emph{frame-label-only} means that spatial annotations from the dataset under study are not used to train the classifier or readout, select checkpoints, or calibrate the detection threshold. Patch locations for copy-paste augmentation are also obtained from model responses, without requiring manually specified target coordinates. The backbone is initialized with weights pretrained for ImageNet classification; we distinguish this external pretraining from supervision supplied by frame labels in the target dataset.

Spatial annotations are used to audit annotation validity, evaluate localization, and diagnose failures. Ground-truth boxes provide the target centers and sizes and the regions used to calculate the signal-to-clutter ratio (SCR). We also train separate baselines with bounding-box supervision. Table~\ref{tab:2-A} distinguishes the annotations needed to run the specified frame-label-only recipe from those used to evaluate and analyze it.

\begin{table*}[tp]
\centering
\caption{\textbf{Annotation use at each stage of the detection pipeline.}}
\label{tab:2-A}
\setlength{\tabcolsep}{4pt}
\renewcommand{\arraystretch}{1.12}
\begin{tabularx}{\textwidth}{@{}>{\hsize=0.87000\hsize\linewidth=\hsize\raggedright\arraybackslash}X>{\hsize=1.29000\hsize\linewidth=\hsize\raggedright\arraybackslash}X>{\hsize=0.84000\hsize\linewidth=\hsize\raggedright\arraybackslash}X@{}}
\toprule
\textbf{Stage} & \textbf{Information used} & \textbf{Spatial annotations from the target dataset} \\
\midrule
Classifier and readout training & Frames, presence/absence labels, and weights pretrained for ImageNet classification & Not used \\
\addlinespace[2pt]
Patch harvesting and synthesis & Training frames, presence/absence labels, model responses, and pixel statistics & Not used \\
\addlinespace[2pt]
Checkpoint selection & Presence/absence labels for separate selection frames & Not used \\
\addlinespace[2pt]
Threshold calibration & Absence labels and map maxima for separate calibration frames & Not used \\
\addlinespace[2pt]
Data audits, localization/size/SCR evaluation, and diagnostics during development & Presence/absence labels and ground-truth boxes & Used \\
\addlinespace[2pt]
Baselines with bounding-box supervision & Bounding boxes from the dataset specified for each comparison and labels for checkpoint selection & Used as specified for each comparison \\
\bottomrule
\end{tabularx}
\end{table*}

\subsection{2.2\quad Model Architecture and Detection Pipeline}

\label{sec:2.2}

\begin{figure*}[tp]
\centering
\includegraphics[width=\textwidth]{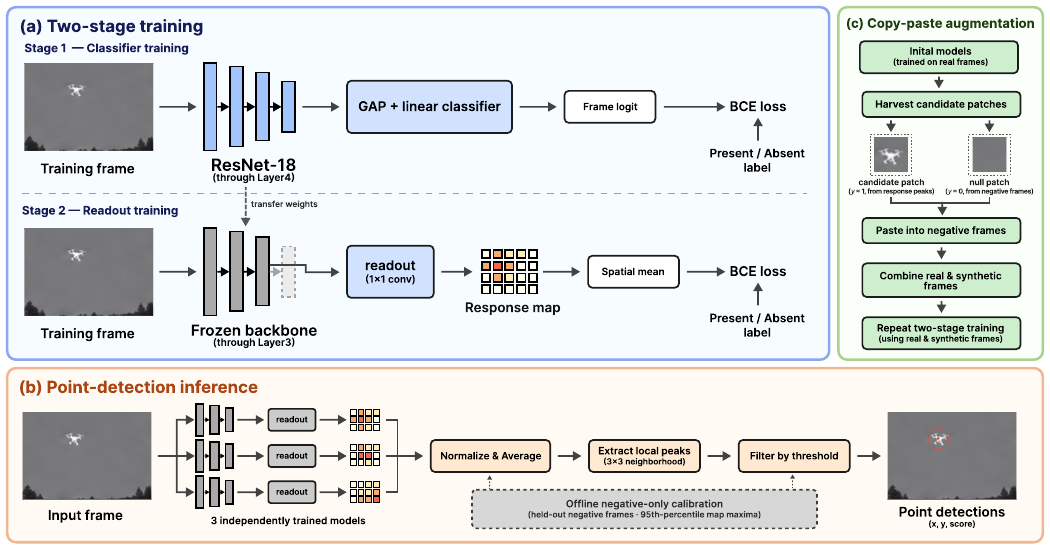}
\caption{\textbf{Frame-label point-detection pipeline: (a) two-stage training; (b) three-model ensemble inference with negative-only calibration; (c) optional copy-paste augmentation and retraining.}}
\label{fig:pipeline}
\end{figure*}

Figure~\ref{fig:pipeline} summarizes two-stage training, ensemble inference, and optional copy-paste augmentation. The pipeline follows the general approach of learning spatial localization responses from image-level classification labels \cite{zhou2016cam,vardazaryan2018}. It consists of an ImageNet-initialized ResNet-18 classifier \cite{he2016} and a linear readout of its intermediate features. For both thermal datasets, we replicate each original 640\ensuremath{\times}512 grayscale frame across three channels and normalize it using the mean and standard deviation computed from frames assigned to training. We retain the original resolution rather than reducing inputs to 224\ensuremath{\times}224, thereby preserving pixel information for small targets. Downsampling in the later ResNet-18 stages is replaced with dilation to maintain an output stride of 8 \cite{yu2017}.

We first train the backbone and a linear classification head using binary cross-entropy (BCE), with global average pooling (GAP) applied to the final features (Figure~\ref{fig:pipeline}(a)). We then freeze the backbone and train a 1\ensuremath{\times}1 convolutional readout on the 256-channel layer3 features. Let $F_3(x;\theta)$ denote these features and $w,b$ the readout weights and bias. The response map is

\begin{equation}
\begin{gathered}
M(x;u,v)=w^\top F_3(x;\theta)_{u,v}+b.
\end{gathered}
\end{equation}

The readout is trained using BCE computed from the spatially averaged map, $z(x)=|\Omega|^{-1}\sum_{(u,v)\in\Omega}M(x;u,v)$, and the frame label. This is equivalent to training a linear classifier on globally averaged layer3 features. Backbone weights and BatchNorm statistics remain fixed during this stage. Neither stage uses target locations or box sizes as training targets. At inference, the backbone through layer3 and the readout produce a 64\ensuremath{\times}80 response map, without an additional trained detector. Map values rank candidate locations and are not interpreted as calibrated detection probabilities.

We selected layer3 based on preliminary Anti-UAV410 experiments with a single ResNet-18 training seed, using ground-truth boxes to evaluate localization. With the backbone frozen and both layers at output stride 8, layer3 yielded better localization than layer4, whereas layer4 yielded better frame-level classification.

Both training stages use AdamW (weight decay $10^{-4}$), a cosine learning-rate schedule, and horizontal flips with probability 0.5. Each epoch contains 288 batches of 32 frames, balanced between positive and negative labels. Twelve within-sequence positive--negative pairs account for 24 frames per batch, without requiring pixel-aligned backgrounds. The remaining eight frames are sampled independently, four per label, by first choosing a sequence uniformly and then a frame uniformly within that sequence.

The classifier is trained for 24 epochs with initial learning rates of $10^{-4}$ for the backbone and $10^{-3}$ for the classification head; the frozen-backbone readout is trained for 6 epochs at $10^{-3}$. These are the default frame-model settings; changes are specified with each comparison. Both stages select checkpoints using the area under the receiver operating characteristic curve (AUROC) for frame-label classification on separate selection frames (Appendix~\ref{sec:A.1}). Final-epoch variants examine dependence on selection.

The pipeline can use real frames alone. The main CST configuration additionally uses bounding-box-free copy-paste augmentation (Figure~\ref{fig:pipeline}(c); Appendix~\ref{sec:A.3}): responses from models trained on real frames identify patches in positive training frames, which are pasted into negative frames to create synthetic positives. Low-contrast null pastes provide synthetic negatives to prevent paste artifacts from becoming positive cues. We retrain the classifier and readout on the expanded data using the same architecture. Synthesis is training-only, and paste locations are not supplied as spatial supervision. Validation diagnostics on CST and A410 identified both positive and negative target-to-background contrast. Copy-paste contrast checks use magnitudes, while synthesis retains the signed patch residual (Appendix~\ref{sec:A.3}). Performance was not evaluated separately by polarity or under intensity inversion. Section~\ref{sec:4} compares configurations with and without synthesis.

We extract local maxima within 3\ensuremath{\times}3 neighborhoods of the response map as candidate locations and output candidates whose scores exceed the threshold as detections. Horizontal and vertical cell coordinates $(u,v)$ in the stride-8 map correspond to image coordinates $(8(u+0.5),8(v+0.5))$. For a single model, the threshold is the 95th percentile of frame-wise map maxima on a separate set of negative calibration frames. This rule limits the fraction of negative calibration frames with high responses; it does not guarantee a false-alarm rate in new scenes.

We report single-model results from three training seeds separately from ensemble results. To form the ensemble in Figure~\ref{fig:pipeline}(b), we divide each seed's raw response map by its own calibration threshold and then average the normalized maps. The per-model thresholds used for ensemble normalization and patch harvesting were all finite and strictly positive. The ensemble threshold is recalculated from the frame-wise maxima of the averaged calibration maps.

\subsection{2.3\quad Datasets, Splits, and Comparisons}

\label{sec:2.3}

The main analysis uses two thermal infrared datasets: CST Anti-UAV and Anti-UAV410 (A410) \cite{xie2025cst,huang2024}. We train a separate detector on each dataset using its provided presence/absence labels. Both datasets are used to examine detection performance, baselines with bounding-box supervision, backbone and spatial-resolution choices, and the effects of synthetic training data. Cross-dataset transfer is evaluated separately. Detailed comparisons of label budgets and training-video coverage are conducted on CST. In the primary balanced test subsets, the median target-box diagonal is 15.4 px for CST and 26.9 px for A410. Because A410 also contains large targets, we report target-size distributions and evaluation radii alongside the results for both datasets.

We re-split CST using connected components of a video-similarity graph to reduce similar or consecutive videos across splits. Training, validation, and test contain 20, 17, and 18 scene groups (93, 55, and 72 sequences), respectively. We retain sequences containing both labels and randomly sample equal positive and negative counts within each sequence without replacement, matching the smaller class size. Some training scene groups are reserved for negative calibration. Primary test subsets are also balanced within sequence, so their positive fraction does not represent target frequency in deployment; all-frame CST test evaluation checks sensitivity to the evaluation distribution. A410 is split using recording identities and background-correlation groups to separate training, selection, calibration, and evaluation. Appendix~\ref{sec:A.1} gives the grouping criteria. These splits share no frame, sequence, or group IDs, but visual scene independence is not guaranteed.

\begin{table}[tp]
\centering
\caption{\textbf{Default CST/A410 real-frame counts. Train: classifier/readout; Select: frame-label checkpoint selection; Calib.: negatives used; Test: balanced subset.}}
\label{tab:2-C}
\setlength{\tabcolsep}{3pt}
\renewcommand{\arraystretch}{1.12}
\begin{tabularx}{\columnwidth}{@{}>{\hsize=1.12000\hsize\linewidth=\hsize\raggedright\arraybackslash}X>{\hsize=0.97000\hsize\linewidth=\hsize\raggedleft\arraybackslash}X>{\hsize=0.97000\hsize\linewidth=\hsize\raggedleft\arraybackslash}X>{\hsize=0.97000\hsize\linewidth=\hsize\raggedleft\arraybackslash}X>{\hsize=0.97000\hsize\linewidth=\hsize\raggedleft\arraybackslash}X@{}}
\toprule
\textbf{Count} & \textbf{Train} & \textbf{Select} & \textbf{Calib.} & \textbf{Test} \\
\midrule
\multicolumn{5}{@{}p{\columnwidth}@{}}{\textbf{CST}} \\
Sequences & 65 & 34 & 17 & 51 \\
Positive & 15,596 & 7,183 & --- & 10,778 \\
Negative & 15,596 & 7,183 & 2,364 & 10,778 \\
\addlinespace[3pt]
\multicolumn{5}{@{}p{\columnwidth}@{}}{\textbf{A410}} \\
Sequences & 27 & 27 & 15 & 26 \\
Positive & 4,611 & 1,924 & --- & 1,920 \\
Negative & 4,611 & 1,924 & 1,153 & 1,920 \\
\bottomrule
\end{tabularx}
\end{table}

Table~\ref{tab:2-C} lists samples assigned to each role; actual checkpoint-selection usage depends on the rule. Label budgets account separately for training, selection, and calibration. CST scene categories are unevenly distributed: jungle and water are absent from training, whereas cn\_sky occurs only there. Category-specific results are interpreted with these differences in mind.

We additionally use Airborne Object Tracking (AOT) as a supplementary dataset to explore applicability beyond thermal imagery \cite{amazonAOT}. A separate model follows Section~\ref{sec:2.2} using helicopter presence/absence labels derived from the annotations of visible-light grayscale images, without copy-paste augmentation.

The AOT experiment expands the training-and-calibration scene pool from 8 to 500 scenes and uses a fixed set of 130 validation scenes for checkpoint selection and localization analysis, with no overlap with training or calibration. Pool sizes include calibration scenes and therefore differ from training-scene counts; both scene count and total frame count increase. Training steps remain fixed across conditions. Section~\ref{sec:4.2} reports the results, and Appendix~\ref{sec:D.1} gives the sampling, allocation, and training settings.

For bounding-box supervision, we compare head training on the same frozen backbone with additional backbone fine-tuning. The head predicts a localization heatmap and target size; localization is evaluated from the heatmap. Compared with the frame-label readout, both the head architecture and loss change, so annotation type is not isolated.

External baselines are label-free frame differencing, using the absolute difference from the pixel-wise median of the preceding eight frames, and COCO-pretrained YOLO26n/s. YOLO uses its own training recipes, and predicted box centers are evaluated under the same localization criteria. The full-data and sparse-label recipes differ in training budgets and checkpoint selection (Sections~\ref{sec:3} and \ref{sec:4.2}). For transfer, training and calibration datasets are specified separately, including whether destination negatives are used for calibration without retraining. Appendix~\ref{sec:A.4} gives implementation details for the bbox heads and frame-differencing baseline.

\subsection{2.4\quad Evaluation Metrics and Aggregation}

\label{sec:2.4}

\textbf{Localization hit rate.} For each positive frame, we bilinearly upsample the response map to the input resolution and find the location of its maximum. We define $\mathrm{hit}_r$ as the fraction of positive frames in which the Euclidean distance between this location and the ground-truth box center is at most $r$. The primary radius for CST is 16 px, with a stricter 8 px radius used to assess finer localization. This metric uses no detection threshold and evaluates only frames containing a target, so it does not account for false alarms on negative frames. For YOLO, we use the center of the highest-confidence retained box as the representative location; a positive frame with no retained boxes counts as a miss. For both hit rates and Pd--FA evaluation, spatial matching uses center distance; predicted box-size errors and intersection-over-union (IoU) with ground truth are not scored.

For comparisons involving datasets with different target sizes, we report size-scaled radii alongside fixed radii. Given ground-truth box width $w$ and height $h$, let $d=\sqrt{wh}$. We define the more permissive radius as $\max(16,1.5d)$ and the stricter radius as $\max(8,0.75d)$. The primary size-scaled columns for A410 and the fixed-radius columns for CST represent different metrics; their names and radii are specified in the tables. Target size is used only to calculate these evaluation radii and is not supplied at inference.

For AOT, localization is evaluated on positive validation frames using the distance from the map maximum to the nearest annotated helicopter center, with fixed radii of 16 px and 8 px. Many AOT boxes use a default size rather than the actual target extent. We therefore do not apply size-scaled radii or box-based analyses of SCR and target size to this dataset.

\textbf{Detection rate and false alarms.} Pd--FA evaluation on CST and A410 uses the set of local peaks in the native-resolution map rather than the single maximum of the upsampled map. A positive frame is counted as detected if at least one peak has a score greater than threshold $t$ and lies within the allowed radius of the target center. $P_d(t)$ is the fraction of positive frames satisfying this condition. To obtain $FA(t)$, the number of false alarms per frame, we count above-threshold peaks outside the allowed radius in positive frames and all above-threshold peaks in negative frames, then divide by the total number of frames. Multiple peaks within the allowed radius are treated as responses to the same target. For YOLO, these rules use the centers and confidence scores of all boxes retained after confidence filtering and non-maximum suppression (NMS), subject to the 300-box limit in Appendix~\ref{sec:A.1}.

We calculate $P_d@FA=a$ as the detection rate at the lowest threshold satisfying $FA(t)\le a$ on the step curve defined by the observed peak scores, without linear interpolation between thresholds. The primary evaluation budget is 0.5 FA/frame; we also report 0.1 FA/frame for a lower false-alarm level. These values characterize the relationship between scores and false alarms using location annotations from the evaluation data. They are distinct from performance at a threshold selected before deployment. Separately, we apply the threshold fixed by negative calibration to the evaluation data and report the fraction of negative frames with an above-threshold response. This frame-level fraction differs from FA, which counts false-alarm peaks per frame.

\textbf{Analysis by difficulty and target size.} We calculate SCR as $|\Delta|/(\sigma_{\mathrm{bg}}+10^{-6})$, where $\Delta$ is the difference between the mean intensity inside the ground-truth box and the mean surrounding-background intensity, and $\sigma_{\mathrm{bg}}$ is the background standard deviation. Background pixels are taken from a surrounding window, excluding the box and a 2 px margin around it. The window side length is $\max(3\max(w,h),21)$ px. SCR summarizes both target-to-background contrast and background variation. Quartile boundaries are determined on the validation data and applied unchanged to the test data. Size analysis uses box-diagonal intervals of [2, 10), [10, 30), [30, 50), and [50, \ensuremath{\infty}) px, with sample counts reported for each interval. We examine SCR and size jointly because they are not independent variables.

\textbf{Repeated runs and uncertainty.} Unless otherwise specified, training is repeated with three seeds. We distinguish hit rates pooled over frames, equally weighted means of sequence-level hit rates, means across single models, and ensemble results. In the reported paired localization comparisons, differences are calculated within each sequence and then averaged with equal weight per sequence. These estimates may therefore differ from a simple subtraction of two frame-pooled results; we specify the aggregation unit for each estimate and confidence interval. Unless otherwise stated, reported paired comparisons use bootstrap resampling of the same sequences for both methods, with trained models held fixed, to obtain 95\% confidence intervals (CIs). We use 2,000 resamples for localization hit rates and 1,000 for Pd@FA, recalculating the operating point within each resample for Pd@FA. However, sequence-level bootstrap intervals may not fully account for dependence among sequences from the same scene group. Results using other analysis units or confidence levels, such as seed-level comparisons of retraining variability, are identified separately.

\textbf{Inference cost.} Accuracy and latency are matched by inference configuration; single-model and ensemble results are reported separately. The main pipeline is timed at 640\ensuremath{\times}512, batch 1, including map normalization, ensemble averaging, maximum and local-peak extraction, and thresholding; input-file loading and device transfer are excluded. Measurements use an RTX A6000 GPU or a CPU with eight threads. After 100 GPU or 20 CPU warm-up iterations, at least 1,024 repetitions are timed for at least 60 seconds in total; GPU timings are synchronized, and medians are reported. GPU precision is either bf16 for the backbone with fp32 for the head and postprocessing, or fp32 throughout. Other backbone and detector comparisons specify their device, precision, and postprocessing conditions. For the YOLO comparison, both methods used the same fixed input; YOLO used torchvision NMS and box clipping.


\section{Detection Performance and Comparisons}

\label{sec:3}

\subsection{3.1\quad Overall Performance of Detectors}

\label{sec:3.1}

Table~\ref{tab:3-A} compares localization and detection under false-alarm constraints on the balanced CST test subset (Section~\ref{sec:2.3}). The main frame-label configuration uses bbox-free copy-paste augmentation; a baseline without synthesis is also included. Results distinguish the primary three-model ensemble from the mean performance of three separately evaluated single models (Section~\ref{sec:2.2}).

\begin{table*}[tp]
\centering
\caption{\textbf{CST balanced-test performance: ensemble / three-seed single-model mean. YOLO reports single-model means only; frame differencing has one result. Dashes indicate unevaluated or inapplicable entries.}}
\label{tab:3-A}
\setlength{\tabcolsep}{4pt}
\renewcommand{\arraystretch}{1.12}
\begin{tabularx}{\textwidth}{@{}>{\hsize=1.74000\hsize\linewidth=\hsize\raggedright\arraybackslash}X>{\hsize=1.14000\hsize\linewidth=\hsize\raggedright\arraybackslash}X>{\hsize=0.78000\hsize\linewidth=\hsize\raggedright\arraybackslash}X>{\hsize=0.78000\hsize\linewidth=\hsize\raggedright\arraybackslash}X>{\hsize=0.78000\hsize\linewidth=\hsize\raggedright\arraybackslash}X>{\hsize=0.78000\hsize\linewidth=\hsize\raggedright\arraybackslash}X@{}}
\toprule
\textbf{Method} & \textbf{Training annotations (UAV)} & \textbf{hit16} & \textbf{hit8} & \textbf{Pd@FA0.1} & \textbf{Pd@FA0.5} \\
\midrule
Frame labels + synthesis (main) & CST frame labels & 0.566 / 0.548 & 0.519 / 0.490 & 0.340 / 0.315 & 0.482 / 0.467 \\
\addlinespace[2pt]
Frame labels, no synthesis & CST frame labels & 0.520 / 0.506 & 0.425 / 0.390 & 0.350 / 0.316 & 0.483 / 0.466 \\
\addlinespace[2pt]
A410 frame-label model \ensuremath{\rightarrow} CST & A410 frame labels & 0.354 / 0.340 & 0.324 / 0.300 & 0.071 / 0.075 & 0.185 / 0.185 \\
\addlinespace[2pt]
Frame differencing & No training & --- / 0.551 & --- / 0.492 & --- / 0.002 & --- / 0.006 \\
\addlinespace[2pt]
Frozen main backbone + A410 bbox head & CST frame labels + A410 boxes & 0.559 / 0.539 & 0.553 / 0.533 & \textbf{0.443} / 0.429 & 0.553 / 0.537 \\
\addlinespace[2pt]
Frozen main backbone + CST bbox head & CST frame labels + CST boxes & 0.576 / 0.558 & 0.570 / 0.549 & 0.427 / 0.389 & 0.581 / 0.556 \\
\addlinespace[2pt]
Main backbone + bbox head, fine-tuned on CST & CST frame labels + CST boxes & \textbf{0.603} / 0.567 & \textbf{0.595} / 0.557 & 0.364 / 0.298 & \textbf{0.603} / 0.559 \\
\addlinespace[2pt]
YOLO26n: CST & CST boxes & --- / 0.687 & --- / 0.685 & --- / 0.553 & --- / 0.665 \\
\addlinespace[2pt]
YOLO26s: CST & CST boxes & --- / \textbf{0.715} & --- / \textbf{0.713} & --- / \textbf{0.609} & --- / \textbf{0.693} \\
\addlinespace[2pt]
YOLO26n: A410 \ensuremath{\rightarrow} CST & A410 boxes & --- / 0.355 & --- / 0.355 & --- / 0.266 & --- / 0.329 \\
\addlinespace[2pt]
YOLO26s: A410 \ensuremath{\rightarrow} CST & A410 boxes & --- / 0.459 & --- / 0.457 & --- / 0.356 & --- / 0.427 \\
\bottomrule
\end{tabularx}
\end{table*}

Table~\ref{tab:3-A} reports hit rates pooled over positive frames; Pd@FA uses the exact step curve and a 16 px radius (Section~\ref{sec:2.4}). The annotation column excludes external pretraining: ImageNet classification for the frame model and COCO detection for YOLO. Frame differencing uses temporal input without training, but its fixed-threshold evaluation still requires separate negative calibration frames.

The main ensemble achieved hit16 of 0.566 and Pd@FA0.5 of 0.482; individual-model hit16 ranged from 0.536 to 0.565. Spatial responses thus provided point detections without target coordinates as training targets. However, the map maximum lay outside the allowed radius in a substantial fraction of positive frames, showing that frame-classification training alone did not fully resolve localization.

Frame differencing achieved hit16 of 0.551, a point estimate close to that of the main configuration, but its Pd@FA0.5 was 0.006. The sequence-averaged hit16 difference between the main ensemble and frame differencing was +0.035 (95\% CI [\ensuremath{-}0.043, +0.106]), so a localization advantage was not established, whereas the Pd@FA0.5 difference was +0.477 (95\% CI [+0.293, +0.582]). In this comparison, the larger difference lay in separating target responses from background responses across frames using a common threshold, rather than in selecting the strongest location within each positive frame. Localization hit rate alone is therefore insufficient to characterize detection performance.

\begin{figure}[tp]
\centering
\includegraphics[width=\columnwidth]{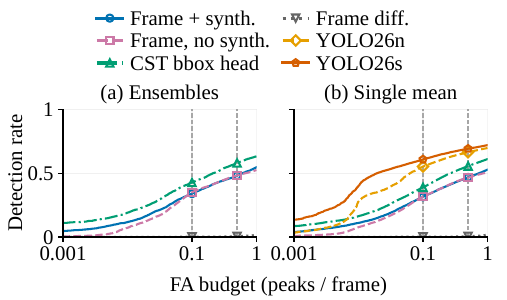}
\caption{\textbf{CST balanced-test Pd--FA (16 px): (a) three-model ensembles; (b) three-seed single-model means at each FA budget. Frame differencing is shared. Markers: 0.1 and 0.5 FA/frame.}}
\label{fig:3-A}
\end{figure}

Adding synthesis increased localization hit rates but left Pd@FA0.5 nearly unchanged, indicating different effects across evaluation criteria (Section~\ref{sec:4.3}). Figure~\ref{fig:3-A} shows the detection--false-alarm tradeoff for the main comparison methods.

\subsection{3.2\quad Comparison with Alternatives Using Bounding-Box Supervision}

\label{sec:3.2}

The bbox comparisons distinguish training a new head on a frozen backbone trained with frame labels, fine-tuning the backbone as well, and using a standard detector (Section~\ref{sec:2.3}). The first two examine changes within the pipeline; the third compares detector systems.

On CST, training a CST bbox head on the frozen main backbone increased ensemble hit16 and hit8 by +0.014 and +0.056 in paired sequence-averaged comparisons. Pd@FA0.1 and Pd@FA0.5 increased by +0.087 and +0.098. The larger gains concerned finer localization and detection at the evaluated FA budgets. Paired hit-rate differences weight sequences equally and can therefore differ from subtracting the frame-pooled table values (Section~\ref{sec:2.4}).

We also examined the same comparison on all CST test frames without balanced sampling: 77,971 frames from 72 sequences, including 67,087 positive and 10,884 negative frames. In this supplementary evaluation, the ensemble Pd@FA0.5 difference pooled over all frames (bbox head \ensuremath{-} main configuration) was +0.127. Both the evaluated sequences and the positive/negative composition differ from those of the balanced test subset, so differences between these evaluations reflect sensitivity to sample composition.

Using A410 boxes to train only the head likewise improved ensemble hit8 and Pd@FA0.5 on CST while leaving hit16 close to the main configuration. Fine-tuning the backbone with CST boxes further increased hit16 relative to training only the CST bbox head. The benefit therefore depended on which components received bbox supervision.

The standard detectors provided a broader comparison. On the same real training frames, YOLO26n/s achieved higher single-model mean hit16 than the main frame-label configuration. YOLO26s exceeded it by +0.143 for sequence-averaged hit16, +0.200 for sequence-averaged hit8, and +0.226 for Pd@FA0.5.

YOLO uses COCO pretraining, its own augmentation and optimizer settings, up to 100 epochs, and checkpoint selection by validation-bbox mAP50--95. These comparisons therefore combine differences in spatial labels, architecture, pretraining, and training and selection procedures. The modest hit16 gain from a bbox head on frozen features and the larger differences relative to standard detectors must be considered together. Section~\ref{sec:4} examines comparisons with fewer labels, and Section~\ref{sec:4.5} compares model size and inference cost.

\subsection{3.3\quad Performance on Anti-UAV410 and Cross-Dataset Transfer}

\label{sec:3.3}

Table~\ref{tab:3-B} reports balanced-test results for models trained on Anti-UAV410 and transferred from CST. The A410 frame-label configuration uses no synthesis; among the rows in Table~\ref{tab:3-B}, copy-paste augmentation was used only during CST training for the ``CST frame-label model \ensuremath{\rightarrow} A410'' row. Because A410 includes larger targets, we report fixed-radius hit rates alongside hit16r and hit8r, using the size-scaled radii defined in Section~\ref{sec:2.4}.

\begin{table*}[tp]
\centering
\caption{\textbf{A410 balanced-test and transfer results, using the reporting convention of Table~\ref{tab:3-A}. Pd@FA0.5 r uses the size-scaled hit16r radius.}}
\label{tab:3-B}
\setlength{\tabcolsep}{4pt}
\renewcommand{\arraystretch}{1.12}
\begin{tabularx}{\textwidth}{@{}>{\hsize=1.80000\hsize\linewidth=\hsize\raggedright\arraybackslash}X>{\hsize=0.84000\hsize\linewidth=\hsize\raggedright\arraybackslash}X>{\hsize=0.84000\hsize\linewidth=\hsize\raggedright\arraybackslash}X>{\hsize=0.84000\hsize\linewidth=\hsize\raggedright\arraybackslash}X>{\hsize=0.84000\hsize\linewidth=\hsize\raggedright\arraybackslash}X>{\hsize=0.84000\hsize\linewidth=\hsize\raggedright\arraybackslash}X@{}}
\toprule
\textbf{Method} & \textbf{hit16r} & \textbf{hit8r} & \textbf{hit16} & \textbf{hit8} & \textbf{Pd@FA0.5 r} \\
\midrule
A410 frame labels & 0.641 / 0.616 & 0.637 / 0.610 & 0.610 / 0.580 & 0.460 / 0.410 & 0.639 / 0.595 \\
\addlinespace[2pt]
Frozen A410 backbone + A410 bbox head & 0.660 / 0.651 & 0.658 / 0.648 & 0.609 / 0.591 & \textbf{0.511} / 0.485 & 0.680 / 0.661 \\
\addlinespace[2pt]
A410 backbone + bbox head, fine-tuned on A410 & \textbf{0.695} / 0.677 & \textbf{0.689} / 0.672 & \textbf{0.645} / 0.625 & \textbf{0.511} / 0.485 & \textbf{0.690} / 0.674 \\
\addlinespace[2pt]
Frozen A410 backbone + CST bbox head & 0.586 / 0.551 & 0.584 / 0.549 & 0.535 / 0.501 & 0.460 / 0.411 & 0.463 / 0.430 \\
\addlinespace[2pt]
CST frame-label model \ensuremath{\rightarrow} A410 & 0.209 / 0.213 & 0.201 / 0.203 & 0.188 / 0.190 & 0.167 / 0.154 & 0.159 / 0.159 \\
\addlinespace[2pt]
YOLO26n: A410 boxes & --- / 0.687 & --- / 0.687 & --- / 0.686 & --- / 0.668 & --- / 0.674 \\
\addlinespace[2pt]
YOLO26s: A410 boxes & --- / \textbf{0.712} & --- / \textbf{0.710} & --- / \textbf{0.709} & --- / \textbf{0.689} & --- / \textbf{0.706} \\
\addlinespace[2pt]
YOLO26n: CST boxes \ensuremath{\rightarrow} A410 & --- / 0.637 & --- / 0.635 & --- / 0.531 & --- / 0.379 & --- / 0.603 \\
\addlinespace[2pt]
YOLO26s: CST boxes \ensuremath{\rightarrow} A410 & --- / 0.615 & --- / 0.611 & --- / 0.498 & --- / 0.371 & --- / 0.568 \\
\bottomrule
\end{tabularx}
\end{table*}

The A410 frame-label ensemble achieved hit16r of 0.641 and Pd@FA0.5 r of 0.639; individual-model hit16r ranged from 0.589 to 0.655. Point detection was thus learned on both datasets using their respective frame labels. Differences in target size, scenes, and training-set size mean that absolute scores across datasets do not establish methodological superiority. Section~\ref{sec:4.3} examines the effects of adding synthesis on A410.

On A410, a bbox-trained head on the frozen backbone increased sequence-averaged ensemble hit16r by +0.006 (95\% CI [\ensuremath{-}0.013, +0.026]), which was not statistically significant. Individual-seed differences were +0.002, +0.056, and +0.032; the latter two had confidence intervals above zero. The ensemble Pd@FA0.5 r difference was +0.041 (95\% CI [+0.005, +0.107]). Bbox-supervision effects therefore depended on the metric and inference configuration. In the single-model comparison, YOLO26s exceeded the frame-label configuration by +0.078 in sequence-averaged hit16r.

Performance of the models trained with frame labels decreased in both transfer directions. The model trained on A410 achieved ensemble hit16 of 0.354 on CST, compared with 0.566 for training on CST itself. Applying the main CST model to A410 yielded ensemble hit16r of 0.209, compared with 0.641 for training on A410 itself. The transfer rows use fixed detector weights, with calibration performed on negative frames from the destination dataset. These results therefore differ from deployment without any use of negative labels from the destination dataset. Section~\ref{sec:5} examines localization failures associated with target size and threshold transfer.

Comparisons with detectors trained on the other dataset also depended on direction and metric. On A410, single-model mean hit16r did not differ significantly between CST-trained YOLO26n/s and the A410-trained frame model. On CST, the locally trained frame model had higher single-model mean hit16 than A410-trained YOLO26s, but its Pd@FA0.5 difference of +0.040 (95\% CI [\ensuremath{-}0.001, +0.085]) was not statistically significant.


\section{Factors Affecting Performance}

\label{sec:4}

Building on the detection results in Section~\ref{sec:3}, we examine how training labels, data composition, features and readouts, and inference configurations affect performance. Diagnostics of the learning process and validation comparisons are used to examine these choices; results verified on test data are identified in the corresponding comparisons.

\subsection{4.1\quad Label Content and Training Stages}

\label{sec:4.1}

Extracting localization responses from CNNs trained with classification labels has been studied in approaches such as CAM \cite{zhou2016cam}. Here, we use the baseline without synthesis to examine how backbone classification training and readout training on frozen features contribute to localization hit rates. We also check spatial responses from simple channel combinations in the main model trained with copy-paste augmentation. All analyses in this subsection use CST validation data. Appendix~\ref{sec:B} details the diagnostic data, training controls, and readout analyses.

\subsubsection{4.1.1\quad Training Labels and Spatial Responses}

\label{sec:4.1.1}

We shuffled presence/absence labels within each training sequence and compared four combinations of true or shuffled labels for the backbone and readout. We used the final checkpoints after 24 epochs of backbone training and 6 epochs of readout training. On the full validation set of 14,366 frames, ensemble hit16 was 0.545 with true labels at both stages, 0.533 when only readout labels were shuffled, 0.009 when only backbone labels were shuffled, and 0.029 when labels were shuffled at both stages. A backbone trained with true labels retained substantial localization performance even when readout labels were shuffled, but individual-model hit16 varied widely, from 0.264 to 0.537.

Subsequent diagnostics used a fixed subset of 3,320 frames, including 1,660 positives. We compared the channel mean with two sets of 32 fixed channel combinations: one with nonnegative weights and the other allowing positive and negative weights. Weights were not selected by localization performance. The channel mean follows an existing approach that aggregates features without learned class-specific weights (LaFAM~\cite{karjauv2024}).

Channel-mean hit16 was 0.002 at ImageNet initialization but ranged from 0.575 to 0.604 across three seeds at the final epoch, indexed as 23, after training with true labels. The range was 0.004--0.013 after training with shuffled labels and 0.003--0.005 for untrained, randomly initialized backbones. We applied the same 32 sets of nonnegative channel weights to each of the three backbones trained with true labels. Across these 96 evaluations, hit16 ranged from 0.504 to 0.614. However, some combinations allowing both positive and negative weights had zero localization hits.

\subsubsection{4.1.2\quad Early Training and BatchNorm Statistics}

\label{sec:4.1.2}

We separately repeated the first 288 batches of the original learning-rate schedule for three seeds, comparing true labels, shuffled labels, and BatchNorm (BN) running-statistics updates alone. Initial parameters and inputs were matched within each seed, with sampling based on true labels. The BN-only control, informed by prior work on adaptation through statistics adjustment \cite{li2016}, kept all parameters at initialization.

\begin{figure}[tp]
\centering
\includegraphics[width=\columnwidth]{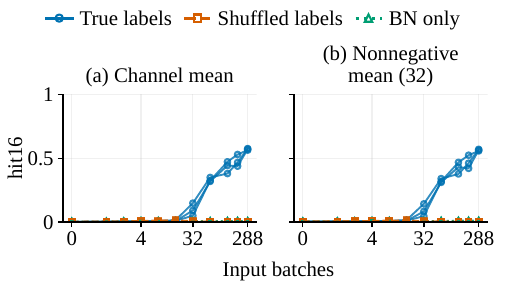}
\caption{\textbf{Early-training hit16 on CST validation data: (a) channel mean; (b) mean over 32 nonnegative combinations. Three seeds per condition; 32 frames/batch.}}
\label{fig:4-A}
\end{figure}

With true labels, all three seeds showed a clear increase in hit16 for the channel mean and nonnegative combinations between the saved checkpoints at 32 and 64 updates, followed by fluctuations. After 288 batches, channel-mean hit16 ranged from 0.5639 to 0.5759 with true labels, from 0 to 0.0006 with shuffled labels, and from 0.0054 to 0.0066 with BN-statistics updates only; nonnegative combinations showed the same ordering (Figure~\ref{fig:4-A}).

Updating BN statistics alone slightly increased hit rates over initialization, but the increase was insufficient to explain the improvement from training with true labels. Resetting only the BN statistics of models trained for 288 updates with true labels reduced channel-mean hit16 to 0.477--0.527, while substantial localization performance remained. This suggests that both learned parameters and the BN statistics used at inference contribute to the observed results.

\subsubsection{4.1.3\quad Readout Stability and Analysis of the Main Model}

\label{sec:4.1.3}

Using three backbones frozen after 24 epochs of true-label training, we repeated readout training with true and shuffled labels under matched initializations and inputs. Readouts trained with true labels achieved high hit rates across all initializations, whereas readouts trained with shuffled labels retained high values only for some combinations (Table~\ref{tab:4-G}).

\begin{table}[tp]
\centering
\caption{\textbf{Readout hit16 ranges on the CST diagnostic subset.}}
\label{tab:4-G}
\setlength{\tabcolsep}{3pt}
\renewcommand{\arraystretch}{1.12}
\begin{tabularx}{\columnwidth}{@{}>{\hsize=0.60000\hsize\linewidth=\hsize\raggedright\arraybackslash}X>{\hsize=1.20000\hsize\linewidth=\hsize\raggedright\arraybackslash}X>{\hsize=1.20000\hsize\linewidth=\hsize\raggedright\arraybackslash}X@{}}
\toprule
\textbf{Backbone seed} & \textbf{True labels (3 readouts)} & \textbf{Shuffled labels (9 readouts)} \\
\midrule
0 & 0.6139--0.6145 & 0.0000--0.6066 \\
\addlinespace[2pt]
1 & 0.6030--0.6060 & 0.0000--0.5861 \\
\addlinespace[2pt]
2 & 0.6108--0.6120 & 0.0000--0.5747 \\
\bottomrule
\end{tabularx}
\end{table}

For each frame, we projected spatially centered features onto the first principal component, the channel combination accounting for the greatest spatial variance. This projection is related to Eigen-CAM~\cite{muhammad2020}; we additionally multiplied it by its signed readout coefficient to isolate its contribution to the learned map. Removing this component from the nine heads trained with true labels reduced hit16 to 0.270--0.464. Among heads trained with shuffled labels, however, one had hit16 of 0.542 for the first component alone but only 0.023 for the full map. Even when a component localizes the target, the final maximum can change depending on its weight relative to other components.

At the selected checkpoints of the main model trained with copy-paste augmentation, channel-mean hit16 ranged from 0.654 to 0.702 across three seeds, and the mean across 32 nonnegative combinations ranged from 0.635 to 0.700. At the final epoch, indexed as 23, however, the learned readout for one seed achieved hit16 of 0.679, compared with 0.594 for the channel mean. Thus, multiple combinations could extract target responses in the main model as well, but differences across combinations and checkpoints remained.

These results suggest that, in this setting, early classification training with true frame labels strengthens target-related spatial responses that can be extracted by multiple nonnegative combinations, and that readout training with true labels helps extract target-related localization responses from frozen features consistently.

\subsection{4.2\quad Label Quantity and Training-Video Coverage}

\label{sec:4.2}

We examined two ways to reduce training labels: retaining only some videos and using their frames, or retaining the coverage of videos while selecting frames sparsely in temporal order within each video. Within each method, we held the number of training steps fixed as labels were reduced. Appendices~\ref{sec:A.1}--\ref{sec:A.2} give the sampling and training-budget rules, including selection and calibration samples.

On CST validation data, the model trained without synthesis achieved ensemble hit16 of 0.531 and hit8 of 0.430 when 2,756 labels were distributed across 65 videos. Three samples that concentrated similar numbers of labels in fewer videos contained 2,760 labels from 8 videos, 2,778 from 7 videos, and 2,754 from 11 videos. Their hit16 values were 0.236, 0.320, and 0.309, respectively, and hit8 ranged from 0.123 to 0.173. All three had lower hit16 than distributing labels across 65 videos, and the 95\% confidence intervals for the sequence-averaged paired differences were all below zero. Even with similar numbers of labeled frames, the videos from which those frames were drawn substantially affected the results.

\begin{figure}[tp]
\centering
\includegraphics[width=\columnwidth]{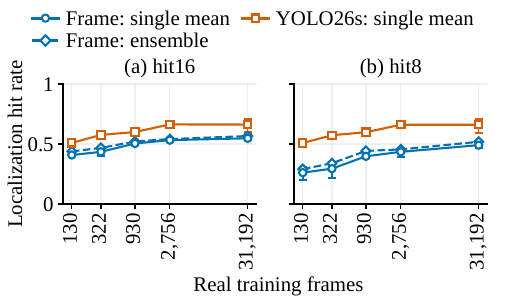}
\caption{\textbf{CST test hit16 (a) and hit8 (b) versus real training-frame count (65 videos). Solid lines: three-seed means; bars: min--max ranges; dashed line: frame-model ensemble.}}
\label{fig:4-B}
\end{figure}

Keeping all 65 videos, we reduced the real-frame budget and generated copy-paste data using only frames and models trained with labels from each budget. Figure~\ref{fig:4-B} compares the frame model and YOLO26s on identical real-frame lists, balanced between positives and negatives. The frame model uses presence/absence labels and copy-paste augmentation; YOLO uses boxes for positives and target-free negatives, so its box count is half the real-frame count. This matches training-frame counts, not annotation effort. Calibration counts differed across frame-model budgets.

YOLO in this subsection uses repeated frames at each budget to maintain the training budget, with checkpoint selection among 40 epochs based on a classification metric using validation frame labels. In Section~\ref{sec:3}, YOLO uses up to 100 epochs and selection based on validation bounding boxes, so even the full-data results in the two sections represent different configurations. The methods also differ in architecture, pretraining, augmentation, and training procedures.

With 31,192 and 2,756 training labels, ensemble hit16 was 0.566 and 0.541, respectively, while Pd@FA0.5 remained similar at 0.482 and 0.480. The difference was larger for hit8 (0.519 versus 0.455), so the apparent preservation of performance depended on localization tolerance. A budget of 930 labels corresponds to approximately 14 per video and yielded a sequence-averaged hit16 difference of \ensuremath{-}0.023 (95\% CI [\ensuremath{-}0.037, \ensuremath{-}0.006]) relative to 2,756 labels. At 322 labels, or approximately 5 per video, the difference from the same reference increased in magnitude to \ensuremath{-}0.079 (95\% CI [\ensuremath{-}0.105, \ensuremath{-}0.055]). These results support sparse labeling under the tested conditions, but performance was not maintained as density was reduced further.

The benefit of retaining training-video coverage was also observed with bounding-box supervision. YOLO26s achieved hit16 of 0.663 with 1,378 bounding boxes distributed across 65 videos, a point estimate close to 0.662 with the full set of bounding boxes. Concentrating a similar number of boxes in 8 videos yielded hit16 of 0.477; the sequence-averaged difference favoring distribution across 65 videos was +0.172. Distributing labels across multiple videos was therefore useful for both training approaches on this dataset, rather than a benefit unique to the frame-label detector.

In addition to training annotations, frame labels were used for checkpoint selection and negative calibration. Checkpoint selection also affected results at low label budgets. For example, with 322 training labels, using the final epoch instead of validation-based selection reduced single-model mean hit16 by 0.009 for the frame model and 0.098 for YOLO. However, validation-based selection of the initial model used to harvest synthetic patches remained in this frame-model comparison, so it did not eliminate selection labels from the entire pipeline.

In the AOT supplementary experiment, the scene pool was expanded from 8 to 500 scenes following Section~\ref{sec:2.3}. For models trained without synthesis, validation ensemble hit16 was 0.452 at 65 scenes and 0.572 at 250 scenes, but 0.490 at 500 scenes. In this comparison, which increased both scene count and total frame count, a larger dataset did not always yield a higher ensemble localization hit rate. Appendix~\ref{sec:D.2} gives the full curves, per-seed results, and additional diagnostics.

Across single-model runs, hit16 ranged from 0.225 to 0.571 at 250 scenes and from 0.025 to 0.575 at 500 scenes. The best single-model hit rates were similar, but the results across the three seeds diverged more at 500 scenes. This observation suggests that expanding data coverage calls for checking both localization performance and stability across training runs.

\subsection{4.3\quad Effects and Applicability of Synthesis Without Bounding-Box Annotations}

\label{sec:4.3}

With the architecture fixed, we compare training with and without the synthesis procedure in Section~\ref{sec:2.2}, examining localization, detection at the evaluated FA budgets, and target-size-specific effects.

On CST test data, synthesis increased ensemble hit16 from 0.520 to 0.566 and hit8 from 0.425 to 0.519. The sequence-averaged paired differences were +0.028 for hit16 and +0.082 for hit8 (95\% CI [+0.061, +0.105]). In contrast, Pd@FA0.5 was 0.483 without synthesis and 0.482 with synthesis, with a difference of \ensuremath{-}0.0002 (95\% CI [\ensuremath{-}0.032, +0.039]). The main benefit of synthesis in this comparison was locating points near the target within positive frames; it did not translate into an improvement in Pd@FA0.5.

Synthesis also increased localization hit rates at low label budgets. On CST validation data, sequence-averaged differences in ensemble hit16 between training with and without synthesis were +0.091, +0.108, and +0.128 at budgets of 930, 322, and 130 labels, respectively. However, the numbers of patches available for synthesis also fell to 194, 51, and 27, and patches could not be obtained from some training-scene categories at the lowest budget. Synthesis can expand target--background combinations obtained from limited labels, but the coverage of its source material also needs to be checked.

Applying the same rules to A410 produced different results. Models without synthesis and models with synthesis using A410 patches were each trained with six seeds. For these A410 synthesis comparisons, 90\% CIs independently resample the six training runs in each condition with evaluation data fixed. The single-model mean difference in validation hit16r was +0.0618 (90\% CI [+0.0457, +0.0780]). When the same models were evaluated on A410 test data, however, the difference was +0.0001 (90\% CI [\ensuremath{-}0.0051, +0.0051]). The validation improvement therefore did not translate into an improvement in overall test hit16r. Table~\ref{tab:4-B} reports this separate set of training runs; these results were not combined with the existing three-seed models in Section~\ref{sec:3}.

\begin{table}[tp]
\centering
\caption{\textbf{A410 test performance without and with copy-paste using A410 patches. Entries are six-seed single-model means; Pd@FA0.5 r uses the hit16r radius.}}
\label{tab:4-B}
\setlength{\tabcolsep}{3pt}
\renewcommand{\arraystretch}{1.12}
\begin{tabularx}{\columnwidth}{@{}>{\hsize=1.08000\hsize\linewidth=\hsize\raggedright\arraybackslash}X>{\hsize=0.96000\hsize\linewidth=\hsize\raggedright\arraybackslash}X>{\hsize=0.96000\hsize\linewidth=\hsize\raggedright\arraybackslash}X@{}}
\toprule
\textbf{Metric} & \textbf{No synthesis} & \textbf{With synthesis} \\
\midrule
hit16r & 0.639 & 0.640 \\
\addlinespace[2pt]
hit8r & 0.630 & 0.635 \\
\addlinespace[2pt]
hit16 & 0.585 & 0.549 \\
\addlinespace[2pt]
hit8 & 0.400 & 0.388 \\
\addlinespace[2pt]
Pd@FA0.5 r & 0.629 & 0.490 \\
\bottomrule
\end{tabularx}
\end{table}

Although overall hit16r changed little, performance improved for some target sizes and declined for others. In Table~\ref{tab:4-C}, hit rates increased for targets with box diagonals of 10--30 px but decreased substantially for those with diagonals of 30--50 px. The two positive frames with box diagonals below 2 px are excluded from Table~\ref{tab:4-C} but included in overall performance. The recipe uses 16\ensuremath{\times}16 px patches, so this fixed patch size and the performance changes across target sizes warrant consideration together. However, without a control varying only patch size, this correspondence does not establish the cause.

\begin{table}[tp]
\centering
\vspace*{0.4pt}
\caption{\textbf{Size-specific A410 test hit16r differences (with \ensuremath{-} without synthesis), using six-seed single-model means and seed-bootstrap 90\% CIs. Size denotes box diagonal; counts are positive frames.}}
\label{tab:4-C}
\setlength{\tabcolsep}{3pt}
\renewcommand{\arraystretch}{1.12}
\begin{tabularx}{\columnwidth}{@{}>{\hsize=0.96000\hsize\linewidth=\hsize\raggedright\arraybackslash}X>{\hsize=0.48000\hsize\linewidth=\hsize\raggedright\arraybackslash}X>{\hsize=0.80000\hsize\linewidth=\hsize\raggedright\arraybackslash}X>{\hsize=1.76000\hsize\linewidth=\hsize\raggedright\arraybackslash}X@{}}
\toprule
\textbf{Size (px)} & \textbf{Count} & \textbf{hit16r diff.} & \textbf{90\% CI} \\
\midrule
2 to <10 & 54 & +0.000 & [\ensuremath{-}0.006, +0.006] \\
\addlinespace[2pt]
10 to <30 & 918 & +0.061 & [+0.050, +0.072] \\
\addlinespace[2pt]
30 to <50 & 164 & \ensuremath{-}0.274 & [\ensuremath{-}0.311, \ensuremath{-}0.232] \\
\addlinespace[2pt]
$\geq50$ & 782 & \ensuremath{-}0.014 & [\ensuremath{-}0.029, \ensuremath{-}0.001] \\
\bottomrule
\end{tabularx}
\end{table}

On A410, the sequence-averaged frame-classification AUROC increased from 0.747 to 0.787, while Pd@FA0.5 r decreased from 0.629 to 0.490. The difference was \ensuremath{-}0.139 (90\% CI [\ensuremath{-}0.193, \ensuremath{-}0.088]). Both hit16 and Pd@FA0.5 also decreased at the fixed 16 px radius. Synthesis therefore cannot be assessed using classification performance or overall localization hit rate alone. Classifier checkpoints in both conditions were selected at early epochs 0--5; the comparison does not establish the effect for models trained longer.

We also examined the option of obtaining patches from another dataset on A410 validation data. Using CST patches for synthesis yielded an overall hit16r difference of +0.0094 relative to no synthesis, but its seed-bootstrap 90\% confidence interval included zero, so an improvement was not established. In the tiny-target interval, however, hit16r was +0.0621 higher than with A410 patches. This interval contained only 59 of the 1,924 positive frames, so the overall mean could obscure such differences. Decisions about synthesis should consider the size ranges of the patches and evaluation targets, alongside dataset identity and overall means.

\subsection{4.4\quad Representations, Readout Placement, and Spatial Resolution}

\label{sec:4.4}

We varied backbone depth and family, initialization, readout placement, and output stride. Each configuration was retrained with three seeds using the classifier, readout, and checkpoint-selection settings in Section~\ref{sec:2.2}. CST used the existing copy-paste data, whereas A410 used real frames only. All results in this subsection are validation comparisons.

\begin{table*}[tp]
\centering
\caption{\textbf{Three-model ensemble validation performance by backbone, stride, and initialization. ImageNet initialization is used unless noted; CST uses synthesis and A410 real frames only.}}
\label{tab:4-H}
\setlength{\tabcolsep}{4pt}
\renewcommand{\arraystretch}{1.12}
\begin{tabularx}{\textwidth}{@{}>{\hsize=2.38000\hsize\linewidth=\hsize\raggedright\arraybackslash}X>{\hsize=0.77000\hsize\linewidth=\hsize\raggedright\arraybackslash}X>{\hsize=0.77000\hsize\linewidth=\hsize\raggedright\arraybackslash}X>{\hsize=0.77000\hsize\linewidth=\hsize\raggedright\arraybackslash}X>{\hsize=0.77000\hsize\linewidth=\hsize\raggedright\arraybackslash}X>{\hsize=0.77000\hsize\linewidth=\hsize\raggedright\arraybackslash}X>{\hsize=0.77000\hsize\linewidth=\hsize\raggedright\arraybackslash}X@{}}
\toprule
\textbf{Configuration} & \textbf{CST hit16} & \textbf{CST hit8} & \textbf{A410 hit16r} & \textbf{A410 hit8r} & \textbf{A410 hit16} & \textbf{A410 hit8} \\
\midrule
ResNet-18, stride 8 & 0.592 & 0.516 & 0.591 & 0.584 & 0.514 & 0.402 \\
\addlinespace[2pt]
ResNet-18, stride 16 & 0.617 & 0.474 & 0.560 & 0.530 & 0.484 & 0.347 \\
\addlinespace[2pt]
ResNet-34, stride 8 & 0.605 & 0.578 & 0.671 & 0.664 & 0.624 & 0.509 \\
\addlinespace[2pt]
ResNet-34, stride 16 & 0.624 & 0.471 & 0.665 & 0.620 & 0.634 & 0.414 \\
\addlinespace[2pt]
MobileNetV3-Small, stride 16 & 0.078 & 0.055 & 0.410 & 0.383 & 0.304 & 0.180 \\
\addlinespace[2pt]
ShuffleNetV2 \ensuremath{\times}1.0, stride 8 & 0.136 & 0.044 & 0.413 & 0.318 & 0.286 & 0.133 \\
\addlinespace[2pt]
ResNet-18, stride 8, random initialization & 0.001 & 0.001 & 0.389 & 0.356 & 0.194 & 0.085 \\
\addlinespace[2pt]
ResNet blocks [1,1,1,1], stride 8, random initialization & 0.007 & 0.006 & 0.287 & 0.259 & 0.160 & 0.113 \\
\bottomrule
\end{tabularx}
\end{table*}

For ResNet, the readout was attached to layer3 to compare strides of 8 and 16. For MobileNetV3-Small \cite{howard2019mobilenetv3} and ShuffleNetV2 \cite{ma2018shufflenetv2}, dilation was adjusted and the readout was attached to the deepest stage at the target stride. The rows for these other families provide contextual comparisons in which both backbone architecture and pretrained representations differ.

Increasing stride from 8 to 16 reduced fixed-radius hit8 for both ResNets on both datasets, even though CST hit16 increased. On A410, size-scaled hit8r also decreased, while hit16 decreased for ResNet-18 and increased slightly for ResNet-34. Performance at 16 px can therefore conceal losses in finer localization; stride 8 was preferable for the two tested ResNets when 8 px precision was required.

Depth interacted with this resolution choice. On CST, ResNet-34 improved ensemble hit8 over ResNet-18 at stride 8, but their hit8 was similar at stride 16. The respective single-seed ranges for ResNet-18 and ResNet-34 were 0.412--0.531 and 0.513--0.571 at stride 8, versus 0.421--0.470 and 0.420--0.446 at stride 16. On A410, ResNet-34 performed better at both strides. Depth and stride should therefore be chosen together, considering the required precision and target dataset.

Frame-classification performance alone could not determine this choice. On CST, MobileNetV3-Small and ShuffleNetV2 achieved sequence-averaged classification AUROC values of 0.740 and 0.733, yet their readouts had low localization hit rates (Table~\ref{tab:4-H}). With the MobileNetV3-Small backbone fixed, moving the readout from the 576-channel output to an earlier 96-channel feature raised ensemble hit16 to 0.135, but the sequence-averaged difference was not statistically significant. This diagnostic indicates sensitivity to readout placement without establishing a reliable recovery or a limitation of the model family as a whole.

Initialization was another important factor: training ResNet-18 at stride 8 from random weights reduced localization performance on both datasets (Table~\ref{tab:4-H}). ImageNet initialization was beneficial under the current training budget. Together with the low hit rates of simple feature readouts before training in Section~\ref{sec:4.1}, this supports starting from pretrained representations and then training with frame labels from the target dataset. Longer training or separate settings for random initialization were not evaluated.

\subsection{4.5\quad Inference Configuration, Latency, and Performance}

\label{sec:4.5}

The deployed frame model omits layer4 and the original classification head because its point-response map is read from layer3. On 64 fixed frames and three seeds, this truncation left readout maps and detection scores unchanged in both bf16 and fp32. Table~\ref{tab:4-D} reports deployment size and performance, with parameter counts including the backbone; the readout adds only 257 parameters per model. The input resolution is 640\ensuremath{\times}512, and floating-point operations (FLOPs) include only convolutions and matrix multiplications. YOLO parameter counts use Conv--BN fusion; the frame model does not. The frame model requires more FLOPs than either YOLO model despite its small readout.

\begin{table}[tp]
\centering
\caption{\textbf{Deployment size and CST test performance (GFLOPs/frame). Three-seed single-model means or three-model ensemble; YOLO uses the full-data Section~\ref{sec:3} recipe.}}
\label{tab:4-D}
\setlength{\tabcolsep}{3pt}
\renewcommand{\arraystretch}{1.12}
\begin{tabularx}{\columnwidth}{@{}>{\hsize=1.70000\hsize\linewidth=\hsize\raggedright\arraybackslash}X>{\hsize=0.95000\hsize\linewidth=\hsize\raggedright\arraybackslash}X>{\hsize=0.95000\hsize\linewidth=\hsize\raggedright\arraybackslash}X>{\hsize=0.70000\hsize\linewidth=\hsize\raggedright\arraybackslash}X>{\hsize=0.70000\hsize\linewidth=\hsize\raggedright\arraybackslash}X@{}}
\toprule
\textbf{Model} & \textbf{Params. (M)} & \textbf{GFLOPs} & \textbf{hit16} & \textbf{hit8} \\
\midrule
Frame, single & 2.783 & 34.4 & 0.548 & 0.490 \\
\addlinespace[2pt]
Frame, ensemble & 8.349 & 103.3 & 0.566 & 0.519 \\
\addlinespace[2pt]
YOLO26n & 2.375 & 4.2 & 0.687 & 0.685 \\
\addlinespace[2pt]
YOLO26s & 9.466 & 16.6 & 0.715 & 0.713 \\
\bottomrule
\end{tabularx}
\end{table}

Under the Section~\ref{sec:2.4} timing protocol, RTX A6000 GPU medians were 2.44 ms for a single truncated model and 6.78 ms for the ensemble, using a bf16 backbone with fp32 head and postprocessing. The ensemble thus cost approximately 2.8 times as much GPU time for the performance gains in Table~\ref{tab:4-D}. All timings below also use 640\ensuremath{\times}512 inputs, batch size 1, and include postprocessing.

Across two fp32 timing runs on a shared server, GPU medians were 1.85--1.86 ms for the truncated frame model, 10.22--10.41 ms for YOLO26n, and 8.34--8.37 ms for YOLO26s. CPU medians on an Intel Xeon Gold 6246R with eight threads reversed the frame-model/YOLO26n ordering (38.9--46.6 versus 33.9--36.8 ms). At batch size 1 in PyTorch's default eager execution mode, latency also depends on kernel-launch overhead, memory traffic, and GPU utilization, which FLOP counts do not capture. The measured YOLO path includes box decoding and NMS, whereas the frame model uses map normalization and local-peak extraction. The relative contributions of these factors were not isolated in this benchmark. The reversed CPU ordering further emphasizes the need to benchmark the intended deployment hardware and inference backend.


\section{Failure Conditions and Remaining Limitations}

\label{sec:5}

The configuration changes examined in Section~\ref{sec:4} improved localization hit rates, but the gains were not uniform across targets and scenes. This section examines remaining failures associated with low SCR, target size and scene composition, cross-dataset transfer, and fixed detection thresholds.

\subsection{5.1\quad Failures Across SCR, Target Sizes, and Scenes}

\label{sec:5.1}

On the CST test set, the main ensemble trained with frame labels achieved hit16 values of 0.301, 0.307, 0.689, and 0.954 from SCR Q1 through Q4, respectively. These intervals contained 2,462, 2,699, 3,126, and 2,491 positive frames. The overall hit16 of 0.566 averages over these differences in difficulty. One positive frame with a box diagonal below 2 px was excluded from Table~\ref{tab:5-A} but included in the overall and SCR-specific results.

\begin{table*}[tp]
\centering
\caption{\textbf{CST test hit16 by SCR and box-diagonal interval, using the main frame-label ensemble. Cells show frame-pooled hit rate (positive count); dashes indicate empty groups.}}
\label{tab:5-A}
\setlength{\tabcolsep}{4pt}
\renewcommand{\arraystretch}{1.12}
\begin{tabularx}{\textwidth}{@{}>{\hsize=1.00000\hsize\linewidth=\hsize\raggedright\arraybackslash}X>{\hsize=1.00000\hsize\linewidth=\hsize\raggedright\arraybackslash}X>{\hsize=1.00000\hsize\linewidth=\hsize\raggedright\arraybackslash}X>{\hsize=1.00000\hsize\linewidth=\hsize\raggedright\arraybackslash}X>{\hsize=1.00000\hsize\linewidth=\hsize\raggedright\arraybackslash}X@{}}
\toprule
\textbf{SCR interval} & \textbf{2 \ensuremath{\leq} diagonal < 10 px} & \textbf{10 \ensuremath{\leq} diagonal < 30 px} & \textbf{30 \ensuremath{\leq} diagonal < 50 px} & \textbf{Diagonal \ensuremath{\geq} 50 px} \\
\midrule
Q1 (low) & 0.201 (378) & 0.295 (1,784) & 0.475 (282) & 0.278 (18) \\
\addlinespace[2pt]
Q2 & 0.142 (339) & 0.346 (1,939) & 0.449 (243) & 0.006 (177) \\
\addlinespace[2pt]
Q3 & 0.300 (353) & 0.738 (2,635) & 0.784 (134) & 0.000 (4) \\
\addlinespace[2pt]
Q4 (high) & 0.989 (1,132) & 0.926 (1,327) & 0.875 (32) & --- \\
\addlinespace[2pt]
All SCR intervals & 0.613 (2,202) & 0.569 (7,685) & 0.544 (691) & 0.030 (199) \\
\bottomrule
\end{tabularx}
\end{table*}

When aggregated by size alone, targets with diagonals below 10 px had hit16 of 0.613, higher than either of the two intermediate size intervals. However, more than half of these targets were in SCR Q4, whereas hit16 for targets of the same size in Q1 and Q2 was 0.201 and 0.142, respectively. Conversely, 177 of the 199 targets with diagonals of at least 50 px were in Q2. Differences between size-group averages therefore cannot readily be attributed to size itself. The fixed 16 px radius is also relatively permissive for small targets.

On the same Q1 subset, single-frame YOLO models achieved higher single-model mean hit16 than the frame-label configuration. This suggests that low-SCR frames retain spatial cues usable by other model and training configurations. Frame differencing also achieved Q1 hit16 of 0.467, but its Pd@FA0.5 on the entire test set was 0.006. Locating targets in difficult positive frames and controlling false alarms when negative frames are also included are distinct aspects of performance. This suggests potential for using temporal cues, with localization and false-alarm control considered together.

\begin{figure*}[tp]
\centering
\includegraphics[width=\textwidth]{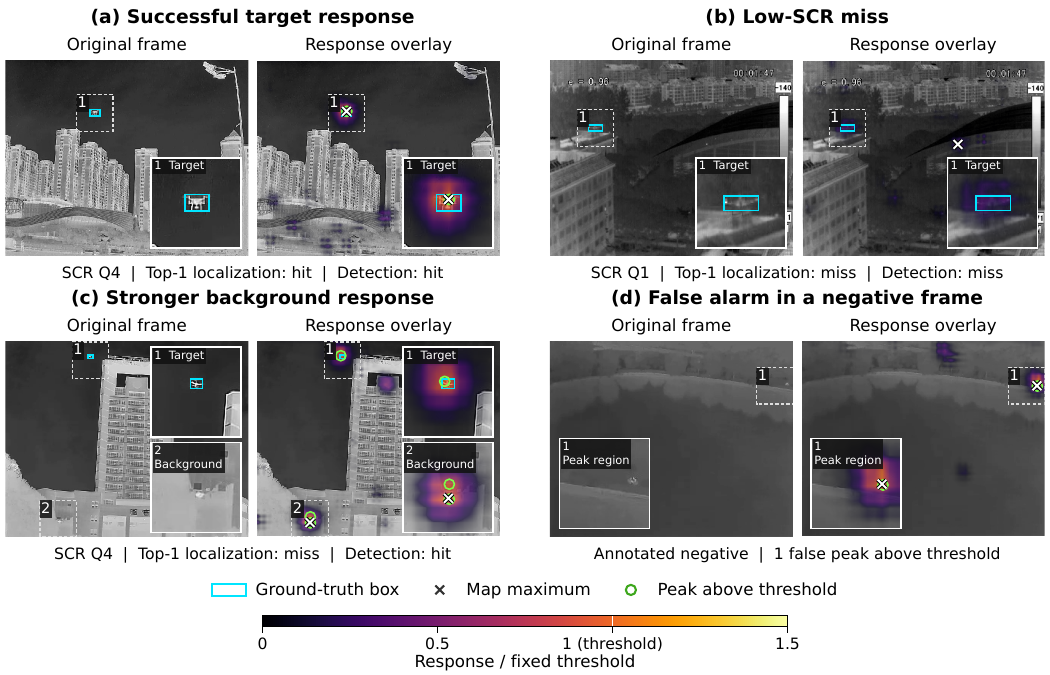}
\caption{\textbf{Main frame-label ensemble outputs on the CST balanced test set \protect\cite{xie2025cst}. Insets show dashed 96\ensuremath{\times}96 px regions. Maps share a scale normalized by the fixed calibration threshold. Localization and detection: Section~\ref{sec:2.4} (16 px).}}
\label{fig:5-A}
\end{figure*}

Performance also varied substantially across scene categories. The main ensemble achieved hit16 of 0.412 in urban-areas (17 sequences, 4,442 positive frames) and 0.995 in jungle (5 sequences, 843 positive frames). In water scenes, comprising 4 sequences and 362 positive frames, hit16 decreased from 0.956 without synthesis to 0.815 with the main synthesis configuration. Neither jungle nor water was represented in the training split, yet their outcomes differed. The absence of a scene category from training alone does not determine its difficulty, and improvements in overall performance should be assessed alongside losses in specific scenes. Figure~\ref{fig:5-A} illustrates successful detection and distinct failure outcomes of the main ensemble.

\subsection{5.2\quad Limitations of Cross-Dataset Transfer and False-Alarm Control}

\label{sec:5.2}

The CST-to-A410 transfer failure reported in Section~\ref{sec:3} was particularly pronounced for large targets. Among 164 A410 positive frames with box diagonals from 30 to less than 50 px, the main frame-label ensemble trained on CST achieved hit16r of 0.012, compared with 0.659 for the ensemble trained with A410 frame labels. For the 782 frames with diagonals of at least 50 px, the corresponding values were 0.074 and 0.951. For the 918 frames with diagonals from 10 to less than 30 px, they were 0.371 and 0.410. The gap for large targets thus remained even when the allowed radius increased with target size. However, the datasets also differ in background and negative-frame composition, making it difficult to explain the observed performance gap solely by a mismatch in target size.

Beyond localization hit rate, score and threshold behavior also changed across datasets. With the main CST frame-label model fixed, replacing CST calibration with A410-negative calibration changed both ensemble-map normalization and the final threshold. On the A410 test set, Pd at the hit16r radius increased from 0.022 to 0.127, FA/frame from 0.013 to 0.203, and the fraction of negative frames with any above-threshold response from 0.006 to 0.097. With A410 calibration, test-curve Pd@FA0.5 r was 0.159; this value does not indicate the detection rate attained by the threshold fixed in advance. Hit16r changed little (0.205 to 0.209), so localization failures remained.

For the frame-label model trained on A410 itself, the threshold determined from 1,153 calibration negatives across 15 sequences produced above-threshold responses in 634 of the 1,920 test negatives (33.0\%). The 95th-percentile calibration rule therefore did not restrict above-threshold responses to approximately 5\% of negative frames in other scenes. This fraction also differs from FA, which counts false-alarm peaks per frame. Recalibrating the threshold with negatives from a new dataset and checking whether it maintains the desired false-alarm level on independent scenes are both necessary.

\subsection{5.3\quad Observed Limitations and Unresolved Causes}

\label{sec:5.3}

The effect of negative-frame composition did not follow a simple relationship. On CST validation data, we compared models trained with several negative-frame compositions favoring easier or harder negatives. Differences in hit16 across configurations were similar in magnitude to the variation observed when retraining the same configuration with different seeds. Small performance differences or the shape of a curve alone therefore did not establish that a particular negative-frame composition was preferable. Confidence intervals obtained by resampling evaluation sequences and variation across training runs represent different sources of uncertainty. Appendix~\ref{sec:C} gives the detailed rules for constructing the training-negative sets and repeating the experiments.

The preceding analyses show that failures span low SCR, target-size and scene distributions, the learning and selection of spatial responses, and the transfer of scores across datasets. The analysis in Section~\ref{sec:4.1} provides evidence that training with true labels and learning the readout contribute to localization responses, but does not isolate a single cause of failure in individual low-SCR frames. These findings are scoped to the evaluated architectures and protocols; their generality across the later WSOL methods discussed in Section~\ref{sec:1.2} has not been evaluated.


\section{Evidence-Based Guidance and Scope of Applicability}

\label{sec:6}

The preceding analyses showed that point detection can be learned from frame labels and that performance depends on data, training, model configuration, and evaluation conditions. This section translates those findings into guidance for choosing data and models and assessing when a frame-label approach may be suitable.

\subsection{6.1\quad Guidance for Configuration Choices and Evaluation}

\label{sec:6.1}

\textbf{Evaluation requirements.} Before choosing training data and model configurations, specify the application's required localization precision and acceptable false-alarm level. Methods with similar localization hit rates showed substantially different Pd at the same FA budget (Section~\ref{sec:3.1}).

\textbf{Label allocation.} When reducing labels, consider sparse frame sampling while retaining training-video coverage, then assess the density needed for the required localization precision. On CST, this allocation yielded higher localization hit rates than concentrating similar label counts in fewer videos for both frame-label and bounding-box training (Section~\ref{sec:4.2}).

\textbf{Learning spatial responses and training the readout.} Pretrained initialization followed by training the backbone and a readout on its frozen features with true frame labels from the target dataset can serve as a baseline (Sections~\ref{sec:4.1} and \ref{sec:4.4}). In the CST controls, classification training strengthened spatial responses, and readouts trained with true labels localized targets more consistently across initializations than those trained with shuffled labels. Simple channel aggregation can diagnose these responses, but its use as a detector requires false-alarm evaluation.

\textbf{Backbone and spatial resolution.} When changing the backbone or readout placement, evaluate localization separately from classification and choose output stride for the required precision. For the two ResNets tested, stride 8 was preferable at an 8 px localization tolerance (Section~\ref{sec:4.4}).

\textbf{Copy-paste augmentation.} Assess synthesis jointly by overall localization hit rate, performance by target size, and Pd at the evaluated FA budgets. CST localization improved without a corresponding gain in Pd@FA0.5, whereas A410 effects depended on size and Pd@FA0.5 r declined (Section~\ref{sec:4.3}). Verify validation gains on separate test data.

\textbf{Difficulty of the target data.} Examine performance and sample counts by target size, SCR, and scene to identify failures hidden by overall averages. On CST, target-size averages also reflected SCR composition, and unseen scene categories differed in performance (Section~\ref{sec:5.1}).

\textbf{False alarms and cross-dataset transfer.} Include negative frames and evaluate detection and false alarms on separate scenes at the threshold fixed before deployment. Calibration on target-dataset negatives did not ensure the intended false-alarm level (Section~\ref{sec:5.2}). Distinguish Pd@FA read from a test curve from performance at a threshold fixed in advance.

\textbf{Stability across training runs and performance differences.} Evaluate changes in configuration or dataset size across multiple training seeds, reporting individual-model variation alongside ensemble performance. Differences in hit16 between negative-frame compositions were comparable to retraining variability, and AOT runs also varied widely (Sections~\ref{sec:4.2} and \ref{sec:5.3}). For small improvements, distinguish uncertainty estimated by resampling evaluation sequences from variability across training seeds.

\textbf{Inference configuration.} Report performance and measured latency together for the single-model or ensemble configuration actually deployed, using the intended hardware. Removing unused layers preserved outputs in our checks, but parameter count and FLOPs alone did not determine execution speed (Section~\ref{sec:4.5}).

\subsection{6.2\quad Annotation Choices and Scope of Applicability}

\label{sec:6.2}

The frame-label approach replaces boundary or center annotations for training with presence/absence judgments from the target dataset, without an external UAV detector. It is an option when frame labels are available or can be obtained by annotating temporal intervals, particularly when spatial annotations are difficult to add. The choice among this approach, full or sparse bounding-box supervision, and transfer of existing detectors depends on annotation convenience, required localization precision, and acceptable false-alarm levels (Sections~\ref{sec:3.3} and \ref{sec:4.2}).

Include labels for checkpoint selection and negative calibration when assessing overall preparation effort; their share of the total label count can grow as training labels decrease. Using the final epoch can reduce dependence on checkpoint selection, although validation-based selection of the initial patch-harvesting model remained in the reported comparison (Section~\ref{sec:4.2}). This pipeline uses frame labels for training and checkpoint selection, while localization evaluation requires reference target locations. Actual annotation time depends on the tools, annotator experience, target visibility, and required label quality. We used the datasets' provided labels and did not measure annotation time, so the matched-frame results do not establish performance under equal annotation effort.

These recommendations draw on the tested models and training conditions on CST and A410, with supplementary AOT analysis. Detailed comparisons of training-video coverage and label density were conducted on CST. AOT used models trained separately with frame labels on visible-light helicopter imagery and evaluated localization and variation across training runs on validation data. This suggests potential beyond thermal imagery; application to new sensors or datasets requires evaluation under their conditions.


\section{Conclusions}

\label{sec:7}

This study analyzed a frame-label UAV point-detection pipeline on CST Anti-UAV and Anti-UAV410, examining its performance, learning behavior, sensitivity to data and model choices, and failure conditions. Given performance gaps relative to bounding-box detectors and the results with sparse bounding-box training, choosing frame labels requires considering annotation convenience alongside required localization precision and acceptable false-alarm levels. Classification or localization gains did not consistently improve detection under false-alarm constraints; evaluation should therefore distinguish these outcomes and account for variation across scenes and training runs.

These analyses provided insights into the architecture's learning and detection behavior and informed guidance for configuring and evaluating frame-label detectors when spatial annotations are difficult to provide. Future work could evaluate other sensors and backbones, compare actual annotation time, and investigate calibration that remains stable across scene changes. We hope that these analyses and observations will help clarify the architecture's potential and characteristics and provide a basis for further architectural improvements and extensions to other application domains.

\FloatBarrier
\addtocontents{toc}{\protect\pagebreak}
\appendices{}
\renewcommand{\theHsection}{appendix.\Alph{section}}

\section{Training and Model Selection Details}

\label{sec:A}

This appendix details checkpoint selection, sampling, and synthesis for Sections~\ref{sec:2} and \ref{sec:4.2}--\ref{sec:4.3}.

\subsection{A.1\quad Checkpoint Selection Rules and Samples}

\label{sec:A.1}

\textbf{Grouping and role separation.} For CST, 16\ensuremath{\times}16 grayscale signatures are normalized to zero mean and unit standard deviation and sampled every five frames. Signature values are scaled by 40, clipped to [\ensuremath{-}127, 127], and stored as integers. Frame-pair distance is the mean absolute difference after rescaling; sequence distance is the smallest mean distance along a temporal offset containing at least 10 sampled frame pairs. Connecting sequences with distance at most 0.15 and taking connected components defines the groups used for splitting. For A410, recording IDs are recovered from sequence names. Twelve evenly spaced frames per recording are resized to 64\ensuremath{\times}64 and combined into a pixel-wise median background; its mean-centered, unit-length signature is grouped by complete linkage at correlation 0.90. Final training, selection, negative-calibration, and test manifests are disjoint in frame, sequence, and group IDs. Synthetic backgrounds and patch sources are restricted to the designated training roles.

Classifier and readout checkpoint selection share time-sorted positive and negative lists within each sequence. Lists of at most 256 frames are used in full; longer lists use 256 evenly spaced positions including both endpoints, rounded to the nearest integer. This yields 9,766 of 14,366 CST frames and 3,752 of 3,848 A410 frames. At each epoch, sequence AUROCs are averaged with equal weight. Table~\ref{tab:2-C} gives counts before sampling.

Classifier selection uses AUROC for the frame-classification logit and AUROC for a score obtained by spatially aggregating the layer4 CAM. For a CAM denoted by $C(x;u,v)$, the second score is

\begin{equation}
\begin{gathered}
s_{\mathrm{LSE}}(x)=\frac{1}{r}\log\left[\frac{1}{|\Omega|}\sum_{(u,v)\in\Omega}\exp\{rC(x;u,v)\}\right],\\[-2pt] r=5.
\end{gathered}
\end{equation}

All selection AUROCs are averaged over sequences, and ties select the earliest epoch. \textbf{Classifier:} retain epochs with logit AUROC within 0.01 of its maximum, then maximize $s_{\mathrm{LSE}}$ AUROC; values within $10^{-4}$ of the best second score are ties. \textbf{Readout:} maximize AUROC of the spatial map mean.

\textbf{Bbox head and fine-tuning:} among 6 epochs, maximize AUROC of the heatmap maximum. \textbf{Sparse-label YOLO:} use the highest box confidence (0 if no detection is returned), selecting the earliest of 40 epochs whose AUROC is within $10^{-4}$ of its maximum.

Sparse-label YOLO uses frames at indices 0, 2, 4, \ldots{} in each sequence-and-label list sorted by time from the same CST selection data, yielding 7,200 frames. Selection inference uses a confidence cutoff of 0.001, NMS IoU of 0.7, and a maximum of 1 detection. Final inference for both YOLO recipes uses the same confidence cutoff and NMS IoU but retains up to 300 detections per frame. Both use fp32 with a 640\ensuremath{\times}512 network input and return boxes in the original image coordinates. By comparison, the standard YOLO models in Section~\ref{sec:3} are trained for up to 100 epochs, and checkpoints are selected using the default detection fitness in Ultralytics 8.4.157 with validation bounding boxes from the respective dataset.

For the final-epoch comparison, the frame-model classifier is fixed at its 24th epoch, and a readout is trained on its frozen features for 6 epochs, with the final readout used for evaluation. Sparse-label YOLO uses its 40th epoch. For frame models using synthesis, the existing patches and synthetic frames are retained. Selection of the initial classifier and readout used for patch harvesting therefore remains validation-based and is not changed by this comparison.

\subsection{A.2\quad Label Budgets and Training-Video Coverage}

\label{sec:A.2}

To reduce the full set of 31,192 real training frames while retaining training-video coverage, the positive and negative lists from each of the 65 training sequences are sorted by time. The first reduced condition retains $m=\max(1,\operatorname{round}(0.08848n))$ frames from a list of length $n$, by rounding $m$ evenly spaced positions including both endpoints and removing duplicate indices. When $m=1$, the first frame is selected. This yields 2,756 real training frames. The same rule is then applied twice, each time retaining approximately one-third of each preceding list, to obtain 930 and 322 frames. The smallest condition selects the middle frame from each sequence-and-label list in the 322-frame set, yielding 130 frames in total. The reduced lists in Section~\ref{sec:4.2} are therefore nested, and the frame models and YOLO share the same real training frames.

In the control that reduces the number of videos, sequence order is shuffled within each scene category using fixed random permutations. A prefix of this order is chosen so that its cumulative label count is closest to 10\% of the category total, retaining at least one sequence per category. Three permutations produced training sets of 8 videos with 2,760 frames, 7 videos with 2,778 frames, and 11 videos with 2,754 frames. These controls were trained without synthesis. Applying the same selection rule within the separate calibration role yielded 272, 340, and 727 negative frames, respectively. For the comparison without synthesis that distributes 2,756 labels across all training videos, calibration uses 272 negative frames selected from 17 videos.

Frame-model step counts and batch size follow Section~\ref{sec:2.2} and remain fixed across label budgets. For sparse-label YOLO with $N$ distinct real frames, each frame is included $\lfloor31{,}192/N\rfloor$ times in the training list. The remaining entries are filled by adding frames once more in filename order, giving a list of 31,192 entries per epoch. Training runs for 40 epochs with this list. These rules maintain the training budget across label budgets within each model family; they do not equalize computation between the frame model and YOLO.

Training-frame counts exclude repeated presentations and synthetic frames. For the label-budget conditions retaining all 65 training sequences, both model families use 2,364 separate negative calibration frames, except for the frame model at 2,756 training frames, which uses 272.

\subsection{A.3\quad Copy-Paste Augmentation Without Bounding Boxes}

\label{sec:A.3}

\textbf{Patch harvesting.} For three models trained on real frames with different seeds, each map is divided by that model's calibration threshold, and the normalized maps are averaged. If the maximum of the averaged map exceeds 1 on a positive training frame, a 16\ensuremath{\times}16 px patch is cropped around the center of the highest-response cell. The patch is rejected if the crop extends outside the image or if the absolute difference between the mean intensities of its central 8\ensuremath{\times}8 region and outer 2 px border is below 0.5. Pasted patches remain centered on the peak cell and are not resized to a smaller scale. At reduced label budgets, patches are harvested only from the real frames in that budget using models trained on those frames.

\textbf{Paste locations and contrast criteria.} For each negative training frame, 8 candidate locations are sampled uniformly from the region excluding a 24 px image border. A nominal 8\ensuremath{\times}8 core is centered at each candidate. Background standard deviation $\sigma$ is calculated from the remaining pixels of a 24\ensuremath{\times}24 window after excluding the core and a surrounding 2 px margin. Pixels in the core are selected by rounding its start and end boundaries down and up, respectively. Eligible locations have candidate $\sigma$ values between the 50th and 95th percentiles within each scene category for CST or each scene group for A410.

Target contrast is set to $\Delta_{\mathrm{target}}=c\sigma_{\mathrm{site}}$. The coefficient $c$ is the 10th percentile of $|\Delta|/(\sigma+10^{-6})$, calculated with the same core and surrounding region on a separate set of recentered crops at the harvested locations. Recentering searches a 32\ensuremath{\times}32 window around each peak for the location with the largest absolute difference between its 3\ensuremath{\times}3 local mean and the window's median intensity. These crops are used to calculate the contrast coefficient; the patches actually pasted are the peak-centered crops defined above.

\textbf{Synthesis operation.} A positive patch is randomly selected from the full harvested pool for the condition. The mean of its outer 2 px border, $\mu_{\partial P}$, is subtracted from patch $P$. The result is multiplied by a Gaussian window $G_u$ with variable width and added to the negative image $I$:

\begin{equation}
\begin{gathered}
I'=I+gG_u(P-\mu_{\partial P}).
\end{gathered}
\end{equation}

We sample $u$ uniformly from 0.3 to 0.6 and set the Gaussian standard deviation along each axis to $8u$ px. Let $\delta_1$ be the added core-to-background contrast when pasting with a gain of 1. We select a patch satisfying $|\delta_1|\geq\max(\Delta_{\mathrm{target}},0.5)$ and apply $g=\min(\Delta_{\mathrm{target}}/|\delta_1|,1)$. Up to 8 patch draws are tried, and the synthesis attempt is skipped if none qualifies. The result is converted to an 8-bit image by rounding and clipping, then saved with JPEG quality 95.

\textbf{Null paste and retraining.} Null pastes use 16\ensuremath{\times}16 crops from negative frames whose absolute mean-intensity difference between the central 8\ensuremath{\times}8 region and outer border is at most 2. The same location-selection, Gaussian-window, and saving procedures are applied, with gain fixed at 1 and the label kept at 0. For each training sequence, the planned counts of synthetic positives and null pastes each equal the number of real positives. Actual counts can be lower when no eligible locations are available or no patch qualifies within the allowed draws. Generated frames are added to the real training data, and input-normalization statistics are recalculated from the combined data before training the classifier and readout.

\textbf{Application across datasets and budgets.} The percentile rule for location selection and the synthesis operation are shared, but the contrast coefficient and eligible locations are recalculated from the pixel statistics of each dataset and budget. In the A410 comparison between patches from A410 and CST, both pools are set to 2,483 patches. The comparison uses the contrast coefficient and candidate locations derived from A410, together with A410 null crops. Locations are drawn separately for each condition. A410 patch harvesting uses models trained without synthesis with seeds 0--2, and the same synthetic data are used to train six models with seeds 0--5. Table~\ref{tab:A-3} reports the actual generated counts rather than planned ratios.

\begin{table}[tp]
\centering
\caption{\textbf{Contrast coefficients and generated-frame counts by synthesis condition. CST row labels give real training-frame counts; synthetic data are shared across seeds.}}
\label{tab:A-3}
\setlength{\tabcolsep}{3pt}
\renewcommand{\arraystretch}{1.12}
\begin{tabularx}{\columnwidth}{@{}>{\hsize=1.60000\hsize\linewidth=\hsize\raggedright\arraybackslash}X>{\hsize=0.52000\hsize\linewidth=\hsize\raggedright\arraybackslash}X>{\hsize=1.00000\hsize\linewidth=\hsize\raggedright\arraybackslash}X>{\hsize=0.88000\hsize\linewidth=\hsize\raggedright\arraybackslash}X@{}}
\toprule
\textbf{Condition} & \textbf{$c$} & \textbf{Synthetic positives} & \textbf{Null pastes} \\
\midrule
CST, 31,192 & 0.466 & 13,361 & 15,596 \\
\addlinespace[2pt]
CST, 2,756 & 0.611 & 1,100 & 1,378 \\
\addlinespace[2pt]
CST, 930 & 0.792 & 332 & 465 \\
\addlinespace[2pt]
CST, 322 & 0.849 & 86 & 157 \\
\addlinespace[2pt]
CST, 130 & 0.457 & 41 & 64 \\
\addlinespace[2pt]
A410 with A410 patches & 0.974 & 3,950 & 4,611 \\
\addlinespace[2pt]
A410 with CST patches & 0.974 & 3,861 & 4,611 \\
\bottomrule
\end{tabularx}
\end{table}

\subsection{A.4\quad Bounding-Box Head and Frame-Differencing Baselines}

\label{sec:A.4}

\textbf{Bounding-box head.} The head applies a 3\ensuremath{\times}3 convolution (256\ensuremath{\rightarrow}128 channels), GroupNorm, and SiLU to layer3 features, followed by two 1\ensuremath{\times}1 convolutions producing localization-heatmap logits and size outputs. For positives, the heatmap target is a Gaussian centered on the bounding-box center; for negatives, it is zero at all locations. The training loss is the heatmap focal loss plus 0.1 times the size L1 loss. The size loss is applied to $\log(w/8)$ and $\log(h/8)$ predicted at the target-center cell in positive frames. The bbox heads in Tables~\ref{tab:3-A} and \ref{tab:3-B} are trained with real frames and bounding boxes from the dataset specified in each row. Although the main CST backbone is trained with synthetic data, synthetic frames are not used to train the bbox head on that backbone.

The head is trained for 6 epochs \ensuremath{\times} 288 steps using AdamW with a head learning rate of $10^{-3}$ and a cosine schedule. In the backbone fine-tuning comparison, the backbone also receives a learning rate of $10^{-4}$, and BN statistics are updated. Checkpoint selection uses frame labels from the designated held-out selection split of CST or A410, following the rules and sampled frames in Appendix~\ref{sec:A.1}. Cross-dataset bbox-head comparisons also use the destination dataset's selection frames; test frames are excluded. Fixed thresholds and ensembles follow the negative-calibration rules in Section~\ref{sec:2.2}. Size outputs are used during training, while the point locations and Pd--FA results in the main text are obtained from the heatmap.

\textbf{Frame differencing.} We take the absolute difference between $I_t$ and the pixel-wise median of the preceding 8 frames, averaging the fourth and fifth sorted values and filling missing initial frames with the first frame. Gaussian smoothing ($\sigma=1$ px) and nonoverlapping 8\ensuremath{\times}8 max pooling produce a 64\ensuremath{\times}80 map. The baseline in Table~\ref{tab:3-A} uses no camera-motion compensation; peak extraction and negative calibration follow Section~\ref{sec:2.2}.


\section{Spatial Response and Readout Diagnostics}

\label{sec:B}

This appendix describes the diagnostics used in Section~\ref{sec:4.1} to analyze the learning of spatial responses and the role of the readout. The analyzed features are the layer3 outputs of ResNet-18 at output stride 8, with spatial dimensions of 64\ensuremath{\times}80 and 256 channels.

\subsection{B.1\quad Diagnostic Data and Corresponding Checkpoints}

\label{sec:B.1}

The four combinations of true or shuffled labels for backbone and readout use the full CST validation set (14,366 frames, 7,183 positives), with BF16 features, an FP32 head, and maps stored in FP16. Subsequent diagnostics use FP32 and a fixed subset: from each time-sorted sequence-and-label list, retain all frames if $n\leq64$; otherwise sample index $\operatorname{round}\{i(n-1)/63\}$ for $i=0,\ldots,63$. This gives 3,320 frames from 34 sequences, with 1,660 per class. The subset is shared across diagnostics, is not chosen using bounding boxes or model success, and is distinct from the checkpoint-selection sample in Appendix~\ref{sec:A.1}.

Epoch indices are zero-based: epochs 0 and 23 follow 288 updates and 24 epochs, respectively. The four label combinations use backbone epoch 23 and readout epoch 5, with three seeds per combination. Feature diagnostics without synthesis compare ImageNet initialization, epochs 0, 1, 2, 5, 11, 17, and 23 of true- or shuffled-label training, and three untrained random initializations.

These diagnostics use normalization from real training data with true labels and no synthesis, disable TF32, and evaluate checkpoints using saved BN statistics unless a reset is specified. ImageNet-initialized features provide a common reference across seeds, distinct from the three random initializations. The main model is also evaluated with its original normalization from training data with synthesis to check the effect of input statistics.

The main model uses three seeds each at the selected and final-epoch states defined in Appendix~\ref{sec:A.1}; the final case pairs backbone epoch 23 with epoch 5 of a newly trained readout. No checkpoint is reselected using localization results. Localization in these diagnostics follows Section~\ref{sec:2.4}, retaining map signs and using bilinear interpolation with \texttt{align\_\allowbreak{}corners=False}; ties select the first maximum in row-major order.

The matched comparisons in \ref{sec:B.3} and \ref{sec:B.4} follow the sampling and horizontal-flip rules in Section~\ref{sec:2.2}. Frame indices and flip decisions are shared across conditions within each matched training or readout seed. Sampling uses true labels, while shuffled loss targets use fixed within-sequence permutations that preserve label counts.

\subsection{B.2\quad Channel Aggregation and Random Combinations}

\label{sec:B.2}

Let $F(p)\in\mathbb{R}^{256}$ denote the features at spatial location $p$ in one frame. The channel-mean map and a map formed with fixed channel weights are

\begin{equation}
\begin{gathered}
M_{\mathrm{mean}}(p)=\frac{1}{256}\sum_{c=1}^{256}F_c(p),\\
M_a(p)=a^\top F(p).
\end{gathered}
\end{equation}

Each weight set uses 32 Gaussian vectors $g\sim\mathcal{N}(0,I_{256})$, converted to FP32 before forming $a=\lvert g\rvert/\lVert g\rVert_2$ for nonnegative combinations or $a=g/\lVert g\rVert_2$ for combinations allowing positive and negative weights. Both sets remain fixed across all frames, backbone seeds, and checkpoints. Neither weights nor map signs are selected from localization results; these combinations use no bias or additional training.

Each combination is scored over all positive frames. Figure~\ref{fig:4-A} tracks the channel-mean hit16 and the arithmetic mean of 32 nonnegative-combination hit16 values for each training seed. Channel-mean ranges in Section~\ref{sec:4.1.2} span the three seeds. The 96-evaluation range in Section~\ref{sec:4.1.1} spans 32 fixed combinations applied to three backbones, not 96 independently trained models.

\subsection{B.3\quad Comparisons During Early Training}

\label{sec:B.3}

The early-training comparisons use 31,192 real frames without synthesis under three conditions: classification with true labels, classification with shuffled labels, and BN-statistics updates only. For each of seeds 0, 1, and 2, the conditions share the same ImageNet-initialized backbone and initial classification head. One label permutation is shared across batches and the three training seeds. Because sampling follows true labels, shuffled loss targets need not be balanced within a batch.

The two classification-training conditions update the backbone, BN affine parameters, and classification head for 288 steps using the AdamW settings in Section~\ref{sec:2.2}. The cosine learning-rate schedule retains its original length of 24\ensuremath{\times}288=6,912 updates, and only its first 288 updates are executed. The BN-statistics-only condition keeps all parameters fixed at initialization and processes the same 288 batches in training mode without gradient computation or optimizer updates. All three conditions update BN running statistics, with momentum 0.1 and epsilon $10^{-5}$, and use BF16 autocast for forward passes. BCE in the training conditions is calculated from FP32 logits.

States are saved after 0, 1, 2, 4, 8, 16, 32, 64, 128, 192, and 288 input batches and evaluated offline under \ref{sec:B.1}, without interleaved validation. Figure~\ref{fig:4-A} counts input batches, including BN-only passes without optimizer updates. These matched-input runs are separate from the existing 24-epoch trajectories described in \ref{sec:B.1}.

An additional evaluation after 288 updates resets the BN running mean, variance, and batch counter of the true-label and shuffled-label models to their initial values, retaining all learned parameters, including BN affine parameters. This replaces statistics after training; it does not train with fixed BN statistics.

\subsection{B.4\quad Repeated Readout Training and Spatial Component Decomposition}

\label{sec:B.4}

\textbf{Repetitions and training.} Three backbones trained with true labels and no synthesis are frozen at epoch 23. For each backbone, three readout initializations are trained with true labels and the same three label permutations shared across all backbones, giving 9 true-label and 27 shuffled-label readouts (36 total). The readout is a 1\ensuremath{\times}1 convolution from 256 channels to one, with bias. Each initialization is shared across the four label conditions, using the matched sampling in \ref{sec:B.1}. Backbone parameters and BN statistics remain frozen in evaluation mode.

Training uses cached FP32 layer3 GAP features of original and horizontally flipped inputs. The map-mean BCE and optimization follow Section~\ref{sec:2.2}, using final epoch 5. Table~\ref{tab:4-G} reports separate ranges for the three true-label and nine shuffled-label readouts per backbone.

\textbf{Component decomposition.} The features of one frame are flattened into $X\in\mathbb{R}^{P\times256}$, with spatial locations as rows and channels as columns ($P=64\times80$). Let $\mu$ be the spatial mean of each channel, and center the features as $X_c=X-\mathbf{1}\mu^\top$. If $v_1$ is the first right singular vector of $X_c$, the map formed with learned readout weights $w$ is decomposed as

\begin{equation}
\begin{gathered}
m_{\mathrm{full}}=X_cw,\\
m_{\mathrm{first}}=(X_cv_1)(v_1^\top w),\\
m_{\mathrm{residual}}=X_cw-m_{\mathrm{first}}.
\end{gathered}
\end{equation}

For each frame, $v_1$ is the unit eigenvector corresponding to the largest eigenvalue of $X_c^\top X_c$; its sign does not affect $m_{\mathrm{first}}$. The signs and magnitudes of learned readout weights are retained, with no absolute-value operation or ReLU on the maps.

The original map $Xw+b\mathbf{1}$ differs from $m_{\mathrm{full}}$ by the spatial constant $(\mu^\top w+b)\mathbf{1}$, so centering preserves its maximum location. For all 36 readouts, the three maps are reshaped to 64\ensuremath{\times}80 and evaluated under \ref{sec:B.1}.


\section{Training-Negative Composition and Variation Across Training Runs}

\label{sec:C}

\textbf{Score used for selection.} The negative-composition comparison in Section~\ref{sec:5.3} ranks training negatives using \texttt{ca\_\allowbreak{}cfar\_\allowbreak{}max}, a local-contrast score calculated from the original images rather than trained-model outputs. Let $J$ be an image with intensities scaled to 0--1. For a location $p$, let $\mu_k(p),\sigma_k(p)$ be the mean and standard deviation of the background pixels in a $3k\times3k$ window centered at $p$, excluding the inner $k\times k$ region. For CST, the window sizes $k\in\{11,17,23\}$ are fixed based on a reference size of 10 px. The frame score is

\begin{equation}
\begin{gathered}
s(J)=\max_{p\in\Omega_{\mathrm{valid}},\,k\in\{11,17,23\}}
\frac{\lvert J(p)-\mu_k(p)\rvert}{\sigma_k(p)+10^{-3}}.
\end{gathered}
\end{equation}

This expression represents the implementation that calculates bright and dark responses separately and takes the larger value. Background statistics are calculated using only valid image pixels within the window, and the search for the maximum excludes a 23 px image border. Lower-scoring frames are called easier and higher-scoring frames harder; these terms refer to the ranking under this local-contrast criterion. Frames in the actual training manifest are scored, and scores saved to five decimal places are used to sort negatives within each sequence. Ties retain the original list order.

\textbf{Sample construction.} All 15,596 real positive frames from the 65 sequences are retained. For a sequence with $n$ negative frames and retained fraction $f$, we select $k=\max(1,\operatorname{round}(fn))$ frames. Rounding is to the nearest integer, with exact ties rounded to the even integer. Easier subsets retain the first $k$ frames of the score-sorted list, and harder subsets retain the last $k$. The selected list is repeated cyclically, and its first $n$ entries are taken to restore the original negative-list length for that sequence.

The seven ranked conditions retain, within each sequence, the easier 25\%, 50\%, and 75\%, all negatives, or the harder 75\%, 50\%, and 25\%. Retaining 25\%, 50\%, 75\%, or 100\% yields 3,901, 7,795, 11,695, or 15,596 distinct negatives, respectively. All conditions contain 15,596 negative-list entries after repetition, before synthesis.

The random-half control randomly selects 50\% of each sequence's negatives without replacement (7,795 distinct frames) and applies the same repetition rule. This subset is fixed once and is not redrawn for each training seed. Only the 100\% condition uses all distinct negatives, so the additional control distinguishes list length from the number of distinct frames.

\textbf{Synthesis and training.} Copy-paste follows Appendix~\ref{sec:A.3}, holding the positive-patch and null-crop pools and the full-data CST contrast coefficient and scene-category background ranges fixed across conditions. Paste locations are restricted to each condition's retained real negatives. One synthetic set is generated using a condition-specific random seed and shared across training seeds: 13,349--13,434 positives and 15,596 null pastes. Thus, matched real-frame list lengths do not imply identical totals after synthesis. Input-normalization statistics are recalculated from each condition's final training data.

Training follows Section~\ref{sec:2.2}; checkpoint selection follows Appendix~\ref{sec:A.1}, with selection and calibration data held fixed. Each condition uses seeds 0, 1, and 2, evaluated individually and as a three-model ensemble on all 14,366 CST validation frames. Localization hit rates use the 7,183 positives.

\textbf{Summarizing variation.} Let $h_{j,s}$ denote the frame-level hit16 of a single model for condition $j$ and training seed $s$ among the seven conditions. Between-condition variation is the sample standard deviation of the seven condition means $\bar h_j$, each averaged over three seeds, and equals 0.0162. Within-condition variation is calculated by taking the sample variance across the three seeds for each condition, averaging the seven variances, and taking the square root, giving 0.0169. These summaries are distinct from variation between ensembles.

Additional models with seeds 3, 4, and 5 for the conditions retaining all negatives and the hardest 25\% within each sequence reuse each condition's data, input statistics, and training and selection rules. Seeds 0--2 and 3--5 form separate ensembles. Differences between conditions or repetitions use 90\% confidence intervals from 2,000 paired resamples of evaluation sequences with trained models held fixed. These intervals quantify evaluation-sample uncertainty; retraining variation is reported separately through the additional seeds.


\section{Settings and Full Results of the AOT Supplementary Experiment}

\label{sec:D}

\subsection{D.1\quad Data Construction and Training Settings}

\label{sec:D.1}

\textbf{Data and sampling.} AOT targets are helicopters in visible-light grayscale imagery \cite{amazonAOT}; a scene is a flight identifier (\texttt{flight\_\allowbreak{}id}). We sort and shuffle 1,801 helicopter-containing flights that each appear in only one distribution part, using seed 20260917. After reserving 100 test and 100 validation scenes, the next 500 form the training/calibration pool. Validation adds 30 helicopter-free scenes. Negatives contain no helicopters but may contain other aircraft.

For a scene containing $L$ original frames sorted by time, we select 50 frames at zero-based indices $\lfloor jL/50\rfloor$ for $j=0,\ldots,49$. The first 8, 18, 35, 65, 125, 250, and 500 scenes in the candidate pool define the comparison conditions. Calibration uses scenes at zero-based positions 9, 19, 29, etc.; if fewer than two are selected, we add scenes from the end of the list. The overall scene pools are therefore nested, but membership in the training and calibration roles is not fully nested in transitions involving the two smallest conditions.

Training sets ranged from 300 frames (133 positive, 167 negative) in 6 scenes to 22,500 frames (8,791 positive, 13,709 negative) in 450 scenes; calibration used 45--1,545 negative frames. Calibration positives are unused; training negatives come from helicopter-containing training scenes. All conditions share 130 validation scenes (6,500 frames: 1,872 positive, 4,628 negative). Validation frame labels select checkpoints; positive frames assess localization.

\textbf{Inputs and training.} Original 2448\ensuremath{\times}2048 images are replicated across three channels and standardized using each condition's training-pixel mean and standard deviation after scaling to 0--1, without resizing, cropping, or synthesis. Architecture, losses, and training settings follow Section~\ref{sec:2.2}, with epochs and steps fixed across conditions. The eight frames outside within-scene pairs are sampled by label. Gradients accumulate over two 16-frame micro-batches, so classifier BatchNorm sees 16 frames. Each condition uses seeds 0, 1, and 2; fixed steps imply changing frame reuse as data grow.

\textbf{Selection and evaluation.} Selection follows Appendix~\ref{sec:A.1} using all 50 frames per scene; scene-level AUROC requires both labels. Calibration and ensembling follow Section~\ref{sec:2.2}. Localization uses fixed 16 px and 8 px radii around the nearest annotated helicopter center (Section~\ref{sec:2.4}).

\subsection{D.2\quad Results and Selection Diagnostics}

\label{sec:D.2}

\begin{figure}[tp]
\centering
\includegraphics[width=\columnwidth]{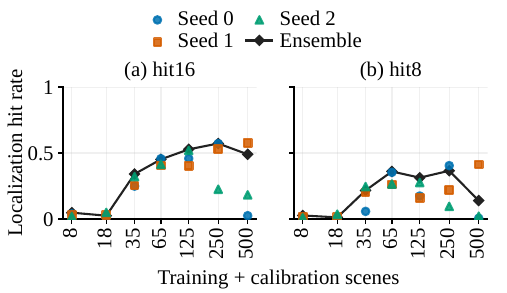}
\caption{\textbf{AOT validation hit16 (a) and hit8 (b) versus training-and-calibration scene count. Colored points show individual training runs, not CIs; the black line shows their three-model ensemble.}}
\label{fig:D-1}
\end{figure}

\textbf{Localization across scene pools.} Figure~\ref{fig:D-1} reports individual-model and ensemble hit16 and hit8 on the same 1,872 validation positives for all seven conditions. Increasing the scene pool did not consistently improve hit16 or hit8. From 250 to 500 scenes, ensemble hit16 decreased and the single-model range widened (Section~\ref{sec:4.2}).

\textbf{Classification and localization.} Across three seeds, the maximum mean scene-level classifier AUROC over 24 epochs increased from 0.784--0.793 at 250 scenes to 0.838--0.846 at 500 scenes. These maxima are distinct from selected-readout performance.

\textbf{Post hoc selection diagnostic.} For the 500-scene condition, a post hoc diagnostic reselected readouts from saved checkpoints using map maxima rather than spatial means, with the same validation frame labels and mean scene-level AUROC criterion. After recalibration, ensemble hit16 increased from 0.490 to 0.521 but remained below 0.572 at 250 scenes; Figure~\ref{fig:D-1} retains the original selection results.

\section*{Acknowledgements}
\addcontentsline{toc}{section}{Acknowledgements}
We acknowledge the providers of CST Anti-UAV~\cite{xie2025cst}, Anti-UAV410~\cite{huang2024}, and Airborne Object Tracking~\cite{amazonAOT} for the datasets used in this study. OpenAI Codex assisted with drafting and revision of the abstract, Sections~\ref{sec:1}--\ref{sec:7}, and Appendices~\ref{sec:A}--\ref{sec:D}, as well as reference checking and plotting-code preparation. OpenAI's image-generation tool was used to create the illustrative UAV images in Figure~\ref{fig:pipeline}. We reviewed the resulting content and take responsibility for the manuscript.

\bibliographystyle{IEEEtran}
\bibliography{references}
\end{document}